\documentclass{article}

\usepackage[final]{corl_2017} 
\usepackage{blindtext}
\usepackage{tikz}
\usepackage{footmisc}

\usepackage{times}
\usepackage{epsfig}
\usepackage{graphicx}
\usepackage{float}
\usepackage{wrapfig}
\usepackage{amsmath,amssymb,amsthm}
\usepackage{algorithm,algorithmicx,algpseudocode}
\usepackage{bm,xspace}
\usepackage{comment}
\usepackage{verbatim}
\usepackage{multirow}
\usepackage{balance}
\usepackage{url}
\usepackage{booktabs}
\usepackage{etoolbox,siunitx}
\usepackage{calc}
\usepackage{pifont,hologo}
\usepackage{color}

\newcommand{\ra}[1]{\renewcommand{\arraystretch}{#1}}

\usepackage[super]{nth}
\usepackage{nicefrac}
\robustify\bfseries

\def\aa{\mathbf{a}}

\def\cc{\mathbf{c}}

\def\oo{\mathbf{o}}

\def\cC{\mathcal{C}}
\def\dD{\mathcal{D}}

\def\Re{\mathbb{R}}

\DeclareMathSymbol{@}{\mathord}{letters}{"3B}

\newcommand\tuple[1]{\left\langle#1\right\rangle}

\newcommand\timess{\mathbin{\!\times\!}}

\newcommand\vladlen[1]{\textcolor{blue}{Vladlen says: #1}}
\definecolor{alexey}{rgb}{0.8,0.,0.8}
\newcommand\alexey[1]{\textcolor{alexey}{Alexey says: #1}}

\newcommand\mypara[1]{\noindent\textbf{#1}}

\def\latex/{\LaTeX}
\def\bibtex/{\hologo{BibTeX}}

\newcommand{\ie}{{\emph{i.e.}}}
\newcommand{\eg}{{\emph{e.g.}}}

\newcommand{\TODO}[1]{{\color{green}[TODO #1]}}
\newcommand{\imitation}{\hspace{0mm}{\fontsize{8}{10}\selectfont Imitation}\hspace{0mm}}
\newcommand{\traditional}{\hspace{0mm}{\fontsize{8}{10}\selectfont Modular}\hspace{0mm}}

\graphicspath{{./images_compressed/}}

\title{CARLA: An Open Urban Driving Simulator}

\author{\!\!Alexey Dosovitskiy\textsuperscript{\normalfont{1}}\!,
German Ros\textsuperscript{\normalfont{2,3}}\!,\,
Felipe Codevilla\textsuperscript{\normalfont{1,3}}\!,
Antonio L\'{o}pez\textsuperscript{\normalfont{3}}\!,
and Vladlen Koltun\textsuperscript{\normalfont{1}}\\[2mm]
\textsuperscript{1}Intel Labs \qquad \textsuperscript{2}Toyota Research Institute \qquad \textsuperscript{3}Computer Vision Center, Barcelona}

\begin{document}
\maketitle

\setlength{\abovedisplayskip}{-2mm}
\setlength{\belowdisplayskip}{-2mm}

\begin{abstract}
We introduce CARLA, an open-source simulator for autonomous driving research. CARLA has been developed from the ground up to support development, training, and validation of autonomous urban driving systems. In addition to open-source code and protocols, CARLA provides open digital assets (urban layouts, buildings, vehicles) that were created for this purpose and can be used freely. The simulation platform supports flexible specification of sensor suites and environmental conditions. We use CARLA to study the performance of three approaches to autonomous driving: a classic modular pipeline, an end-to-end model trained via imitation learning, and an end-to-end model trained via reinforcement learning. The approaches are evaluated in controlled scenarios of increasing difficulty, and their performance is examined via metrics provided by CARLA, illustrating the platform's utility for autonomous driving research.
\end{abstract}



\section{Introduction}
\label{sec:introduction}

Sensorimotor control in three-dimensional environments remains a major challenge in machine learning and robotics. The development of autonomous ground vehicles is a long-studied instantiation of this problem~\cite{Pomerleau1988,Silver2010navigation}. Its most difficult form is navigation in densely populated urban environments~\citep{Paden2016}. This setting is particularly challenging due to complex multi-agent dynamics at traffic intersections; the necessity to track and respond to the motion of tens or hundreds of other actors that may be in view at any given time; prescriptive traffic rules that necessitate recognizing street signs, street lights, and road markings and distinguishing between multiple types of other vehicles; the long tail of rare events -- road construction, a child running onto the road, an accident ahead, a rogue driver barreling on the wrong side; and the necessity to rapidly reconcile conflicting objectives, such as applying appropriate deceleration when an absent-minded pedestrian strays onto the road ahead but another car is rapidly approaching from behind and may rear-end if one brakes too hard.

Research in autonomous urban driving is hindered by infrastructure costs and the logistical difficulties of training and testing systems in the physical world. Instrumenting and operating even one robotic car requires significant funds and manpower. And a single vehicle is far from sufficient for collecting the requisite data that cover the multitude of corner cases that must be processed for both training and validation. This is true for classic modular pipelines~\cite{Paden2016,Franke2017} and even more so for data-hungry deep learning techniques. Training and validation of sensorimotor control models for urban driving in the physical world is beyond the reach of most research groups.

An alternative is to train and validate driving strategies in simulation. Simulation can democratize research in autonomous urban driving. It is also necessary for system verification, since some scenarios are too dangerous to be staged in the physical world (e.g., a child running onto the road ahead of the car). Simulation has been used for training driving models since the early days of autonomous driving research~\cite{Pomerleau1988}. More recently, racing simulators have been used to evaluate new approaches to autonomous driving~\cite{Wymann2014torcs,Chen:2015}. Custom simulation setups are commonly used to train and benchmark robotic vision systems~\cite{Biedermann2016,Gaidon:2016,Haltakov:2013,Handa2012,Mueller2016,Ros:2016,Skinner2016,Zhang2016:ICRA}. And commercial games have been used to acquire high-fidelity data for training and benchmarking visual perception systems~\cite{Richter:2016,Richter:2017}.

While ad-hoc use of simulation in autonomous driving research is widespread, existing simulation platforms are limited. Open-source racing simulators such as TORCS~\cite{Wymann2014torcs} do not present the complexity of urban driving: they lack pedestrians, intersections, cross traffic, traffic rules, and other complications that distinguish urban driving from track racing. And commercial games that simulate urban environments at high fidelity, such as Grand Theft Auto V~\cite{Richter:2016,Richter:2017}, do not support detailed benchmarking of driving policies: they have little customization and control over the environment, limited scripting and scenario specification, severely limited sensor suite specification, no detailed feedback upon violation of traffic rules, and other limitations due to their closed-source commercial nature and fundamentally different objectives during their development.

\begin{figure}[t]
        \centering
        \setlength{\tabcolsep}{0.5mm}
        \begin{tabular}{@{}cc@{}}
          \includegraphics[width=0.495\textwidth]{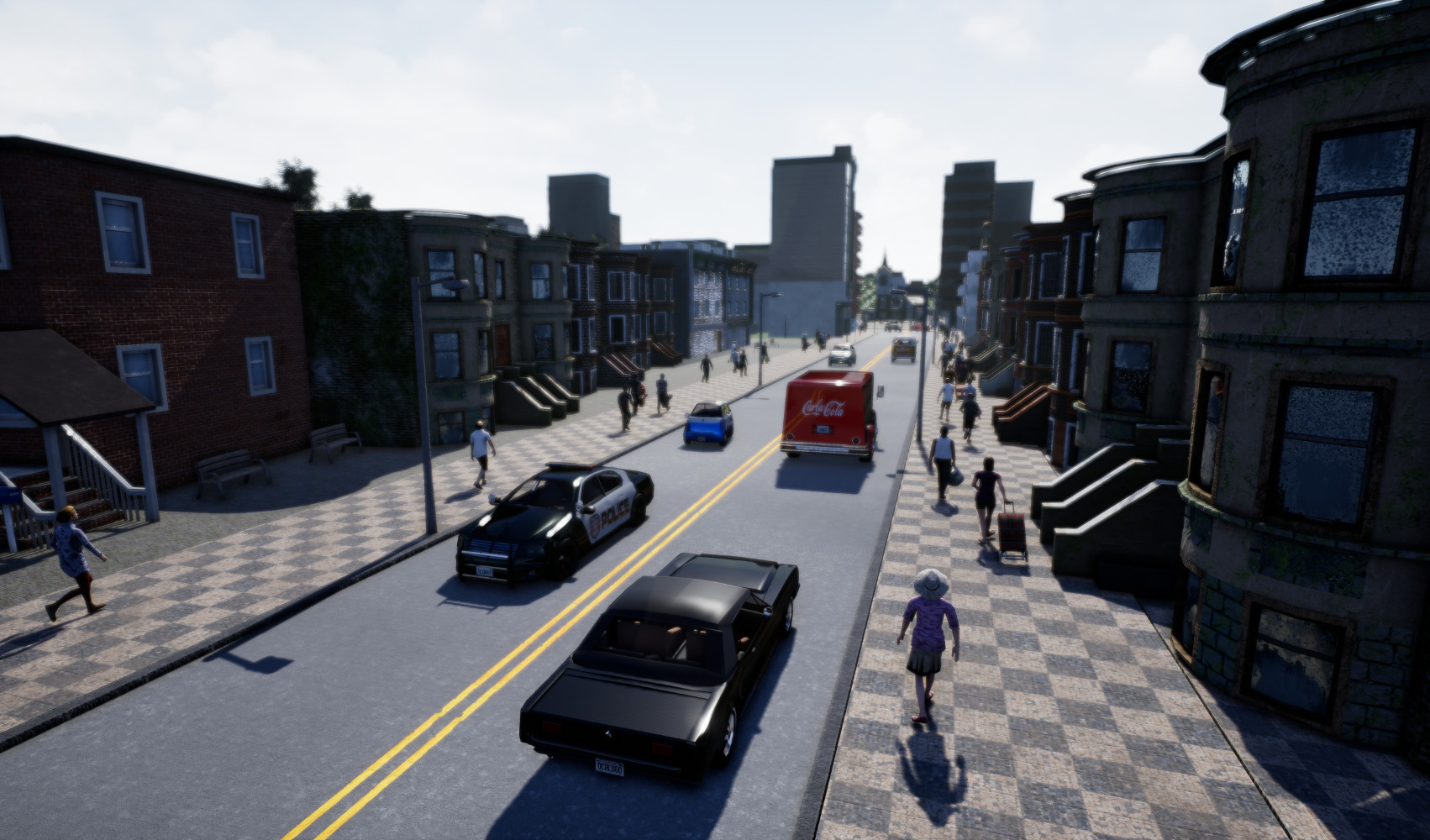} &
          \includegraphics[width=0.495\textwidth]{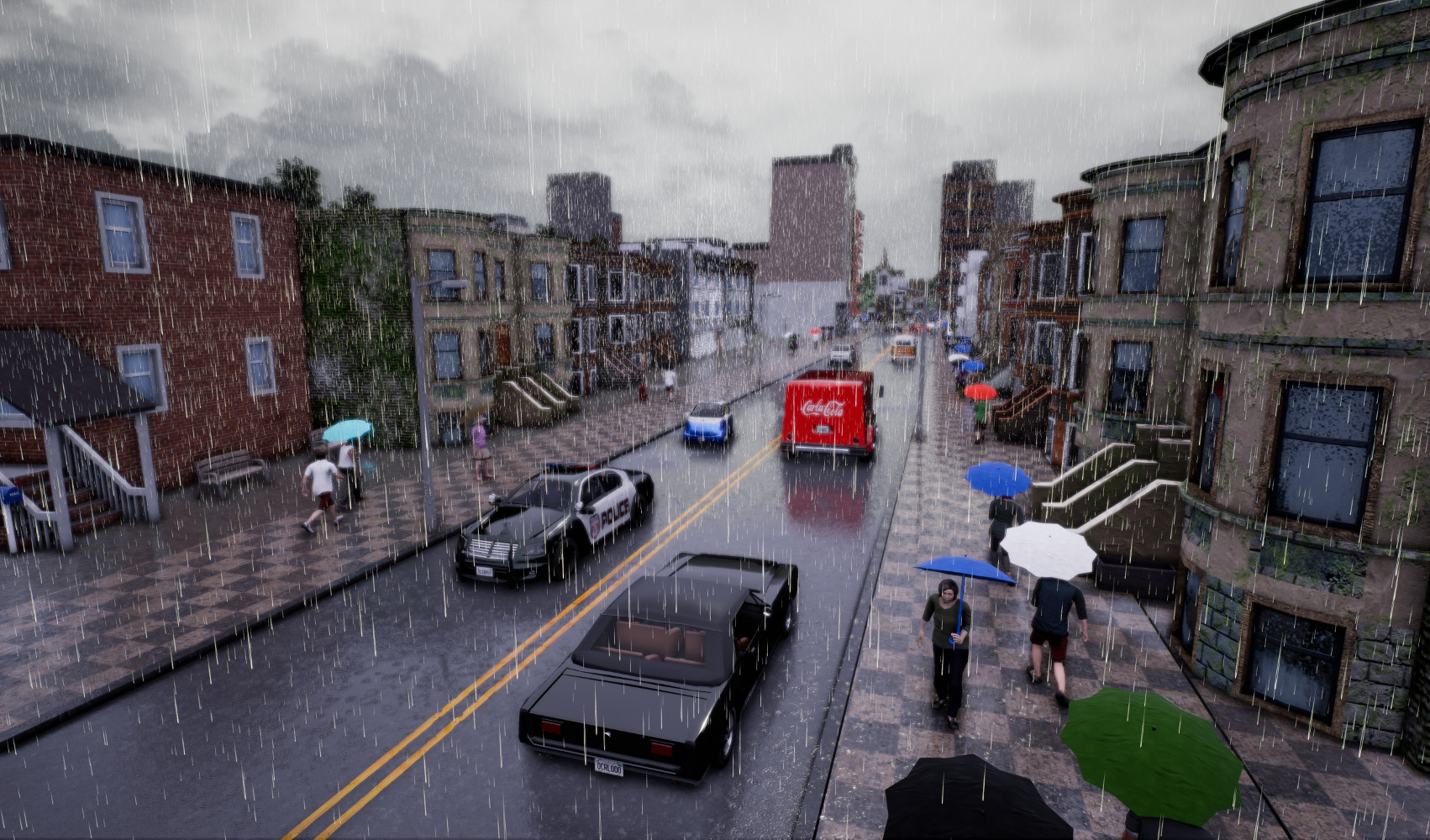} \\
          \includegraphics[width=0.495\textwidth]{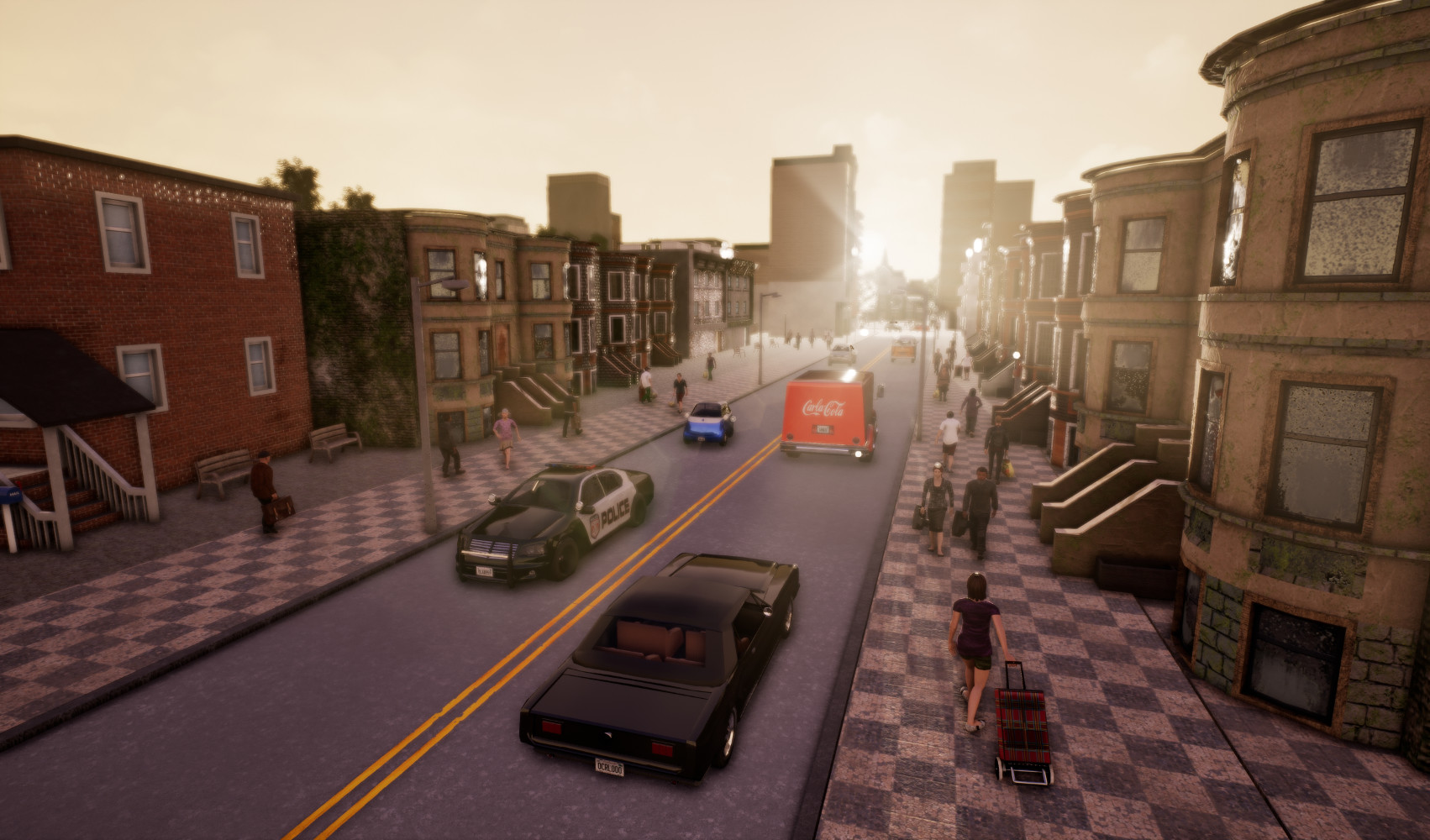} &
          \includegraphics[width=0.495\textwidth]{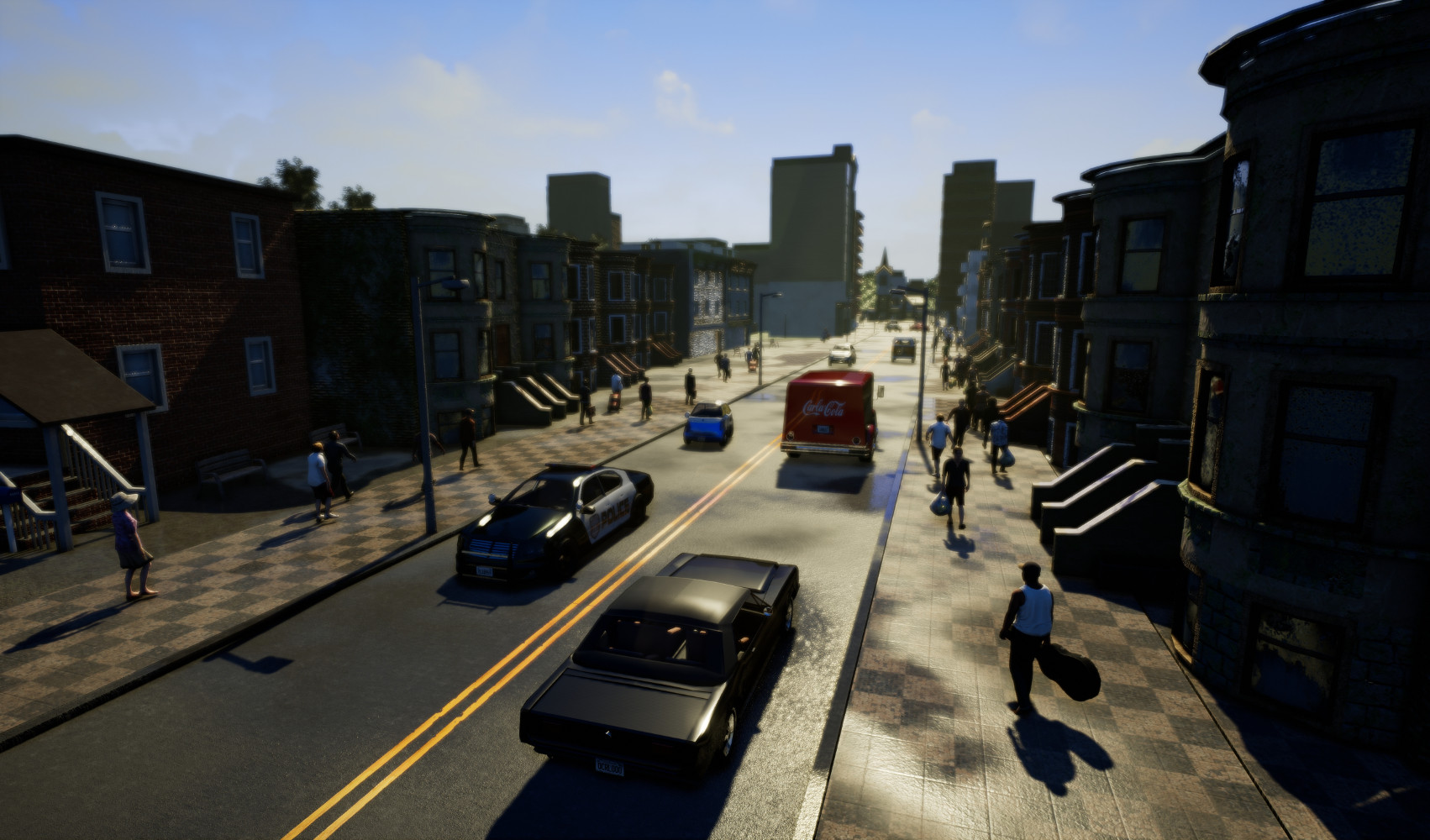}
        \end{tabular}
        \vspace{-2mm}
\caption{A street in Town 2, shown from a third-person view in four weather conditions. Clockwise from top left: clear day, daytime rain, daytime shortly after rain, and clear sunset. See the supplementary video for recordings from the simulator.}
        \label{fig:street}
        \vspace{-3mm}
\end{figure}

In this paper, we introduce CARLA (Car Learning to Act) -- an open simulator for urban driving. CARLA has been developed from the ground up to support training, prototyping, and validation of autonomous driving models, including both perception and control. CARLA is an open platform.
Uniquely, the content of urban environments provided with CARLA is also free.
The content was created from scratch by a dedicated team of digital artists employed for this purpose. It includes urban layouts, a multitude of vehicle models, buildings, pedestrians, street signs, etc. The simulation platform supports flexible setup of sensor suites and provides signals that can be used to train driving strategies, such as GPS coordinates, speed, acceleration, and detailed data on collisions and other infractions. A wide range of environmental conditions can be specified, including weather and time of day. A number of such environmental conditions are illustrated in Figure~\ref{fig:street}.

We use CARLA to study the performance of three approaches to autonomous driving. The first is a classic modular pipeline that comprises a vision-based perception module, a rule-based planner, and a maneuver controller.
The second is a deep network that maps sensory input to driving commands, trained end-to-end via imitation learning.
The third is also a deep network, trained end-to-end via reinforcement learning.
We use CARLA to stage controlled goal-directed navigation scenarios of increasing difficulty.
We manipulate the complexity of the route that must be traversed, the presence of traffic, and the environmental conditions.
The experimental results shed light on the performance characteristics of the three approaches.

\section{Simulation Engine}
\label{sec:simulator}

CARLA has been built for flexibility and realism in the rendering and physics simulation.
It is implemented as an open-source layer over Unreal Engine 4 (UE4)~\cite{UE4}, enabling future extensions by the community. The engine provides state-of-the-art rendering quality, realistic physics, basic NPC logic, and an ecosystem of interoperable plugins. The engine itself is free for non-commercial use.

CARLA simulates a dynamic world and provides a simple interface between the world and an agent that interacts with the world. To support this functionality, CARLA is designed as a server-client system, where the server runs the simulation and renders the scene. The client API is implemented in Python and is responsible for the interaction between the autonomous agent and the server via sockets. The client sends commands and meta-commands to the server and receives sensor readings in return. Commands control the vehicle and include steering, accelerating, and braking. Meta-commands control the behavior of the server and are used for resetting the simulation, changing the properties of the environment, and modifying the sensor suite. Environmental properties include weather conditions, illumination, and density of cars and pedestrians. When the server is reset, the agent is re-initialized at a new location specified by the client.

\mypara{Environment.}
The environment is composed of 3D models of static objects such as buildings, vegetation, traffic signs, and infrastructure, as well as dynamic objects such as vehicles and pedestrians. All models are carefully designed to reconcile visual quality and rendering speed: we use low-weight geometric models and textures, but maintain visual realism by carefully crafting the materials and making use of variable level of detail. All 3D models share a common scale, and their sizes reflect those of real objects. At the time of writing, our asset library includes $40$ different buildings, $16$ animated vehicle models, and $50$ animated pedestrian models.

We used these assets to build urban environments via the following steps:
(a) laying out roads and sidewalks;
(b) manually placing houses, vegetation, terrain, and traffic infrastructure;
and (c) specifying locations where dynamic objects can appear (spawn).
This way we have designed two towns: Town 1 with a total of $2.9$ km of drivable roads, used for training, and Town 2 with $1.4$~km of drivable roads, used for testing.
The two towns are shown in the supplement.

One of the challenges in the development of CARLA was the configuration of the behavior of non-player characters, which is important for realism. We based the non-player vehicles on the standard UE4 vehicle model (PhysXVehicles). Kinematic parameters were adjusted for realism. We also implemented a basic controller that governs non-player vehicle behavior: lane following, respecting traffic lights, speed limits, and decision making at intersections. Vehicles and pedestrians can detect and avoid each other. More advanced non-player vehicle controllers can be integrated in the future~\cite{Best2017}.

Pedestrians navigate the streets according to a town-specific navigation map, which conveys a location-based cost.
This cost is designed to encourage pedestrians to walk along sidewalks and marked road crossings, but allows them to cross roads at any point. Pedestrians wander around town in accordance with this map, avoiding each other and trying to avoid vehicles. If a car collides with a pedestrian, the pedestrian is deleted from the simulation and a new pedestrian is spawned at a different location after a brief time interval.

To increase visual diversity, we randomize the appearance of non-player characters when they are added to the simulation.
Each pedestrian is clothed in a random outfit sampled from a pre-specified wardrobe and is optionally equipped with one or more of the following: a smartphone, shopping bags, a guitar case, a suitcase, a rolling bag, or an umbrella. Each vehicle is painted at random according to a model-specific set of materials.

We have also implemented a variety of atmospheric conditions and illumination regimes. These differ in the position and color of the sun, the intensity and color of diffuse sky radiation, as well as ambient occlusion, atmospheric fog, cloudiness, and precipitation. Currently, the simulator supports two lighting conditions~-- midday and sunset~-- as well as nine weather conditions, differing in cloud cover, level of precipitation, and the presence of puddles in the streets. This results in a total of $18$ illumination-weather combinations. (In what follows we refer to these as weather, for brevity.) Four of these are illustrated in Figure~\ref{fig:street}.

\begin{figure}[t]
        \centering
        \setlength{\tabcolsep}{0.5mm}
        \begin{tabular}{@{}ccc@{}}
          \includegraphics[width=0.328\textwidth]{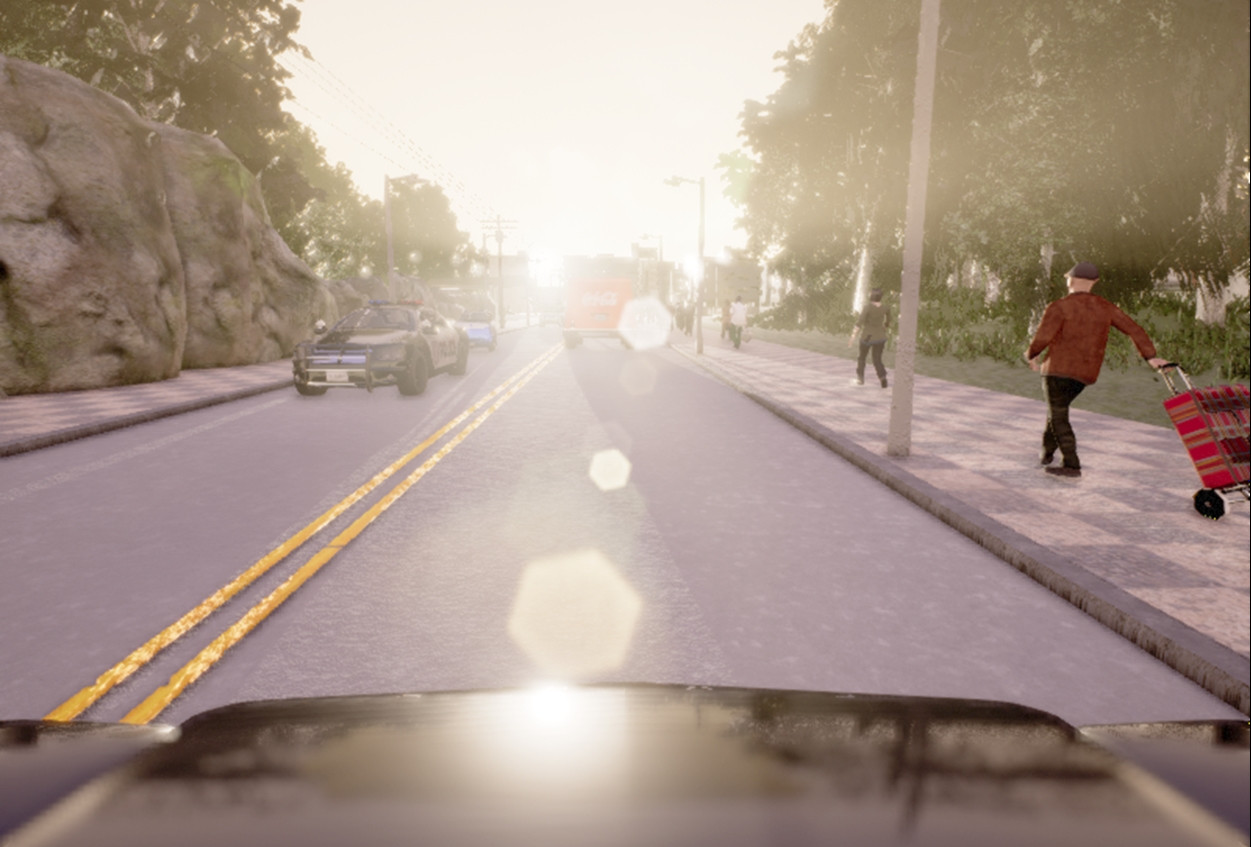} &
          \includegraphics[width=0.328\textwidth]{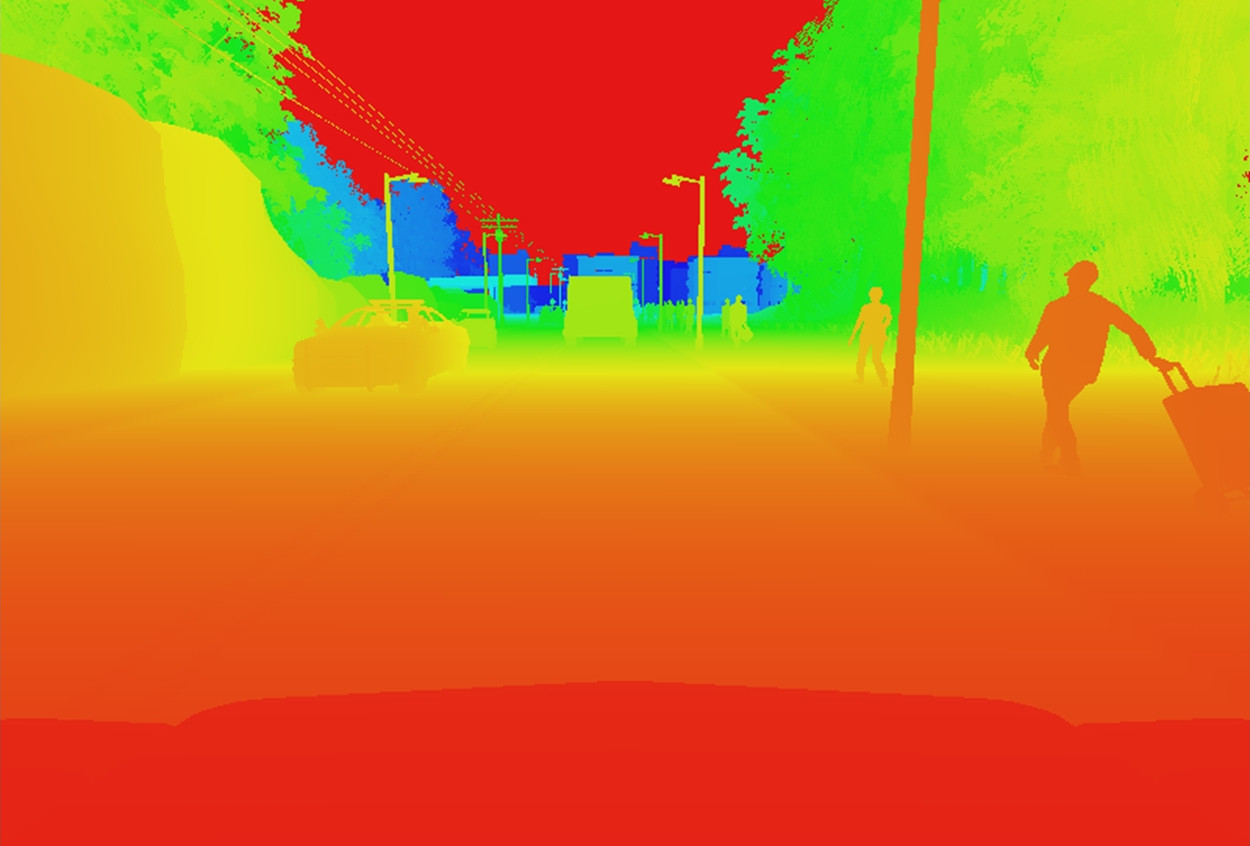} &
          \includegraphics[width=0.328\textwidth]{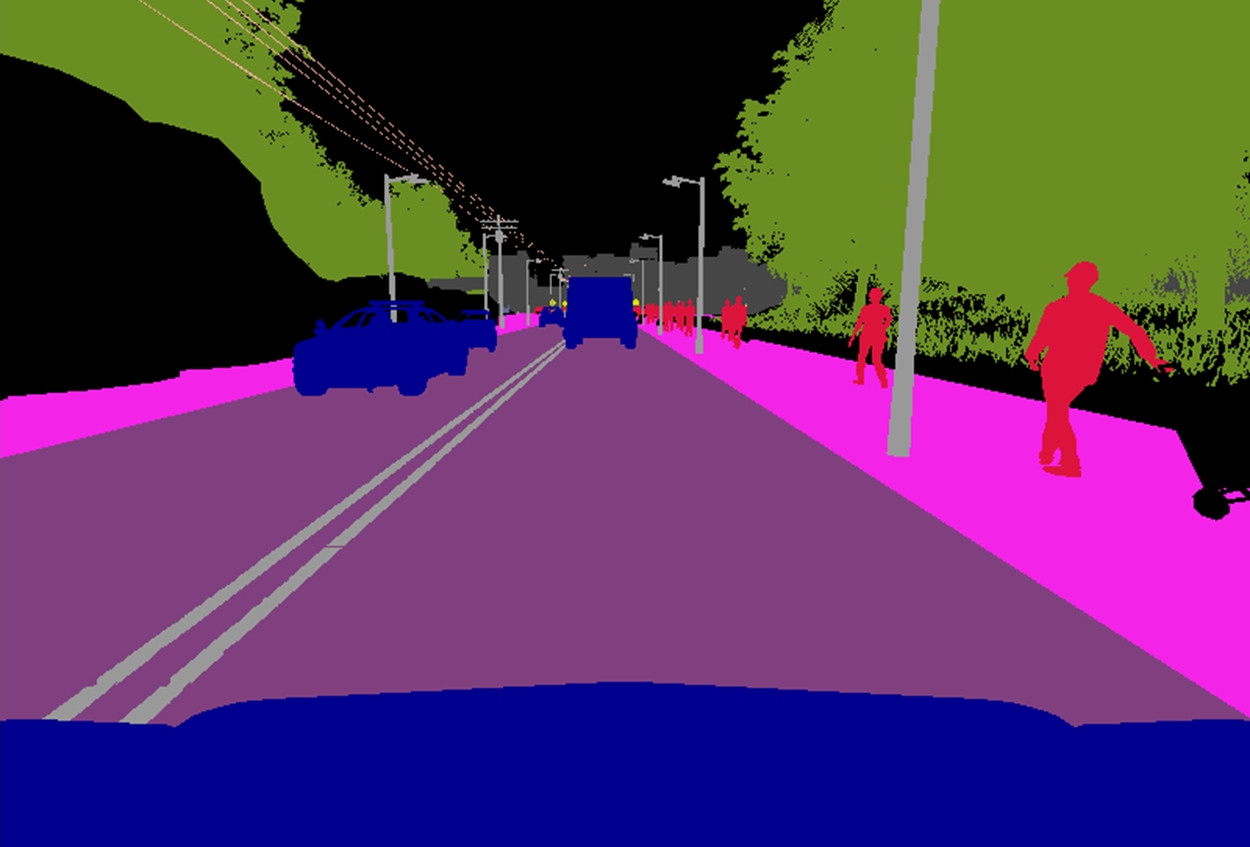}
        \end{tabular}
        \vspace{-2mm}
\caption{Three of the sensing modalities provided by CARLA. From left to right: normal vision camera, ground-truth depth, and ground-truth semantic segmentation. Depth and semantic segmentation are pseudo-sensors that support experiments that control for the role of perception. Additional sensor models can be plugged in via the API.}
        \label{fig:onboard}
        \vspace{-3mm}
\end{figure}

\mypara{Sensors.}
CARLA allows for flexible configuration of the agent's sensor suite. At the time of writing, sensors are limited to RGB cameras and to pseudo-sensors that provide ground-truth depth and semantic segmentation. These are illustrated in Figure~\ref{fig:onboard}. The number of cameras and their type and position can be specified by the client.
Camera parameters include  3D location,  3D orientation with respect to the car's coordinate system, field of view, and depth of field. Our semantic segmentation pseudo-sensor provides 12 semantic classes: road, lane-marking, traffic sign, sidewalk, fence, pole, wall, building, vegetation, vehicle, pedestrian, and other.

In addition to sensor and pseudo-sensor readings, CARLA provides a range of measurements associated with the state of the agent and compliance with traffic rules.
Measurements of the agent's state include vehicle location and orientation with respect to the world coordinate system (akin to GPS and compass), speed, acceleration vector, and accumulated impact from collisions.
Measurements concerning traffic rules include the percentage of the vehicle's footprint that impinges on wrong-way lanes or sidewalks, as well as states of the traffic lights and the speed limit at the current location of the vehicle.
Finally, CARLA provides access to exact locations and bounding boxes of all dynamic objects in the environment.
These signals play an important role in training and evaluating driving policies.

\section{Autonomous Driving}
\label{sec:methods}

CARLA supports development, training, and detailed performance analysis of autonomous driving systems. We have used CARLA to evaluate three approaches to autonomous driving. The first is a modular pipeline that relies on dedicated subsystems for visual perception, planning, and control. This architecture is in line with most existing autonomous driving systems~\cite{Paden2016,Franke2017}.
The second approach is based on a deep network trained end-to-end via imitation learning~\cite{Codevilla2017}. This approach represents a long line of investigation that has recently attracted renewed interest~\cite{Pomerleau1988,LeCun2005driving,Codevilla2017}.
The third approach is based on a deep network trained end-to-end via reinforcement learning~\cite{Mnih2016a3c}.

We begin by introducing notation that is common to all methods and then proceed to describe each in turn.
Consider an agent that interacts with the environment over discrete time steps.
At each time step, the agent gets an observation $\oo_t$ and must produce an action $\aa_t$.
The action is a three-dimensional vector that represents the steering, throttle, and brake.
The observation $\oo_t$ is a tuple of sensory inputs. This can include high-dimensional sensory observations, such as color images and depth maps, and lower-dimensional measurements, such as speed and GPS readings.

In addition to momentary observations, all approaches also make use of a plan provided by a high-level topological planner. This planner takes the current position of the agent and the location of the goal as input, and uses the $A^*$ algorithm to provide a high-level plan that the agent needs to follow in order to reach the goal. This plan advises the agent to turn left, turn right, or keep straight at intersections. The plan does not provide a trajectory and does not contain geometric information. It is thus a weaker form of the plan that is given by common GPS navigation applications which guide human drivers and autonomous vehicles in the physical world. We do not use metric maps.

\subsection{Modular pipeline}

Our first method is a modular pipeline that decomposes the driving task among the following subsystems: (i) perception, (ii) planning, and (iii) continuous control. Since no metric map is provided as input, visual perception becomes a critical task. Local planning is completely dependent on the scene layout estimated by the perception module.

The perception stack uses semantic segmentation to estimate lanes, road limits, and dynamic objects and other hazards. In addition, a classification model is used to determine proximity to intersections. The local planner uses a rule-based state machine that implements simple predefined polices tuned for urban environments. Continuous control is performed by a PID controller that actuates the steering, throttle, and brake. We now describe the modules in more detail.


\mypara{Perception.}
The perception stack we describe here is built upon a semantic segmentation network based on RefineNet~\cite{Lin:2017}.
The network is trained to classify each pixel in the image into one of the following semantic categories:  $\cC =$ \{\texttt{road}, \texttt{sidewalk}, \texttt{lane marking}, \texttt{dynamic object}, \texttt{miscellaneous static}\}.
The network is trained on $2@500$ labelled images produced in the training environment using CARLA.
The probability distributions provided by the network are used to estimate the ego-lane based on the road area and the lane markings.
The network output is also used to compute an obstacle mask that aims to encompass pedestrians, vehicles, and other hazards.

In addition, we estimate the likelihood of being at an intersection by using a binary scene classifier (intersection/no intersection) based on AlexNet~\cite{Krizhevsky:2012}. This network is trained on $500$ images balanced between the two classes. 

\mypara{Local planner.}
The local planner coordinates low-level navigation by generating a set of waypoints: near-term goal states that represent the desired position and orientation of the car in the near future. The goal of the planner is to synthesize waypoints that keep the car on the road and prevent collisions. The local planner is based on a state machine with the following states: (i)~road-following, (ii)~left-turn, (iii)~right-turn, (iv)~intersection-forward, and (v)~hazard-stop. Transitions between states are performed based on estimates provided by the perception module and on topological information provided by the global planner.
Further details can be found in the supplement.
The local plan in the form of waypoints is delivered to the controller, along with the vehicle's current pose and speed.

\mypara{Continuous controller.}
We use a proportional-integral-derivative (PID) controller~\cite{Emirler:2014} due to its simplicity, flexibility, and relative robustness to slow response times. Each controller receives the current pose, speed, and a list of waypoints, and actuates the steering, throttle and brake, respectively. We target a cruise speed of 20 km/h. Controller parameters were tuned in the training town.

\subsection{Imitation learning}
\label{sec:method_imitation}

Our second method is conditional imitation learning, a form of imitation learning that uses high-level commands in addition to perceptual input~\cite{Codevilla2017}.
This method utilizes a dataset of driving traces recorded by human drivers in the training town. The dataset ${\dD= \{ \tuple{\oo_i, \cc_i, \aa_i} \}}$ consists of tuples, each of which contains an observation $\oo_i$, a command $\cc_i$, and an action $\aa_i$.
The commands are provided by drivers during data collection and indicate their intentions, akin to turn signals.
We use a set of four commands: follow the lane (default), drive straight at the next intersection, turn left at the next intersection, and turn right at the next intersection. The observations are images from a forward-facing camera. To increase the robustness of the learned policies, we inject noise during data collection. The dataset is used to train a deep network to predict the expert's action $\aa$ given an observation $\oo$ and a control command $\cc$.
Further details are provided by Codevilla et al.~\cite{Codevilla2017}.

We have collected around $14$ hours of driving data for training.
The network was trained using the Adam optimizer~\cite{Kingma2015adam}.
To improve generalization, we performed data augmentation and dropout.
Further details are provided in the supplement.

\subsection{Reinforcement learning}
\label{sec:method_rl}
Our third method is deep reinforcement learning, which trains a deep network based on a reward signal provided by the environment, with no human driving traces. We use the asynchronous advantage actor-critic (A3C) algorithm~\cite{Mnih2016a3c}. This algorithm has been shown to perform well in simulated three-dimensional environments on tasks such as racing~\cite{Mnih2016a3c} and navigation in three-dimensional mazes~\cite{Mnih2016a3c,Jaderberg2017unreal,Dosovitskiy2017dfp}.
The asynchronous nature of the method enables running multiple simulation threads in parallel, which is important given the high sample complexity of deep reinforcement learning.

We train A3C on goal-directed navigation.
In each training episode the vehicle has to reach a goal, guided by high-level commands from the topological planner.
The episode is terminated when the vehicle reaches the goal, when the vehicle collides with an obstacle, or when a time budget is exhausted.
The reward is a weighted sum of five terms: positively weighted speed and distance traveled towards the goal, and negatively weighted collision damage, overlap with the sidewalk, and overlap with the opposite lane.
Further details are provided in the supplement.


The network was trained with $10$ parallel actor threads, for a total of $10$ million simulation steps.
We limit training to $10$ million simulation steps because of computational costs imposed by the realistic simulation.
This correspond to roughly $12$ days of non-stop driving at $10$ frames per second.
This is considered limited training data by deep reinforcement learning standards, where it is common to train for hundreds of millions of steps~\cite{Mnih2016a3c}, corresponding to months of subjective experience.
To ensure that our setup is fair and that $10$ million simulation steps are sufficient for learning to act in a complex environment, we trained a copy of our A3C agent to navigate in a three-dimensional maze (task D2 from Dosovitskiy and Koltun~\cite{Dosovitskiy2017dfp}).
The agent reached a score of $65$ out of $100$ after $10$ million simulation steps~-- a good result compared to $60$ out of $100$ reported by Dosovitskiy and Koltun~\cite{Dosovitskiy2017dfp} after $50$ million simulation steps for A3C with less optimized hyperparameters.

\section{Experiments}
\label{sec:experiments}



We evaluate the three methods~-- modular pipeline (MP), imitation learning (IL), and reinforcement learning (RL)~-- on four increasingly difficult driving tasks, in each of the two available towns, in six weather conditions.
Note that for each of the three approaches we use the same agent on all four tasks and do not fine-tune separately for each scenario.
The tasks are set up as goal-directed navigation: an agent is initialized somewhere in town and has to reach a destination point.
In these experiments, the agent is allowed to ignore speed limits and traffic lights. We organize the tasks in order of increasing difficulty as follows:
\begin{itemize}
\setlength{\itemsep}{0pt}
\setlength{\parskip}{0pt}
\setlength{\parsep}{0pt}
\item Straight: Destination is straight ahead of the starting point, and there are no dynamic objects in the environment. Average driving distance to the goal is 200 m in Town 1 and 100 m in Town 2.
\item One turn: Destination is one turn away from the starting point; no dynamic objects. Average driving distance to the goal is 400 m in Town 1 and 170 m in Town 2.
\item Navigation: No restriction on the location of the destination point relative to the starting point, no dynamic objects. Average driving distance to the goal is 770 m in Town 1 and 360 m in Town 2.
\item Navigation with dynamic obstacles: Same as the previous task, but with dynamic objects (cars and pedestrians).
\end{itemize}

Experiments are conducted in two towns. Town 1 is used for training, Town 2 for testing.
We consider six weather conditions for the experiments, organized in two groups.
Training Weather Set was used for training and includes clear day, clear sunset, daytime rain, and daytime after rain.
Test Weather Set was never used during training and includes cloudy daytime and soft rain at sunset.

For each combination of a task, a town, and a weather set, testing is carried out over $25$ episodes.
In each episode, the objective is to reach a given goal location.
An episode is considered successful if the agent reaches the goal within a time budget. The time budget is set to the time needed to reach the goal along the optimal path at a speed of $10$ km/h.
Infractions, such as driving on the sidewalk or collisions, do not lead to termination of an episode, but are logged and reported.

\section{Results}
\label{sec:results}

Table~\ref{tbl:completion} reports the percentage of successfully completed episodes under four different conditions.
The first is the training condition: Town 1, Training Weather Set. Note that start and goal locations are different from those used during training: only the general environment and ambient conditions are the same.
The other three experimental conditions test more aggressive generalization: to the previously unseen Town 2 and to previously unencountered weather from the Test Weather Set.

Results presented in Table~\ref{tbl:completion} suggest several general conclusions.
Overall, the performance of all methods is not perfect even on the simplest task of driving in a straight line, and the success rate further declines for more difficult tasks.
Generalization to new weather is easier than generalization to a new town.
The modular pipeline and the agent trained with imitation learning perform on par on most tasks and conditions.
Reinforcement learning underperforms relative to the other two approaches.
We now discuss these four key findings in more detail.

\begin{table}[]
\small
\centering
\ra{1.2}
\resizebox{1.0\linewidth}{!}{
 \begin{tabular}{@{}lcccccccccccccccc@{}}
\toprule

       && \multicolumn{3}{c}{Training conditions} && \multicolumn{3}{c}{New town} && \multicolumn{3}{c}{New weather} && \multicolumn{3}{c}{New town\,\&\,weather} \\
    Task             && MP   & IL   & RL   && MP   & IL   & RL   && MP   & IL   & RL   && \;MP\; & \;IL   & \; RL \;   \\
    \midrule
    Straight         && $\textbf{98}$ & $95$ & $89$ && $92$ & $\textbf{97}$ & $74$ && $\textbf{100}$ & $98$ & $86$ && $50$   & \;$\textbf{80}$ & $68$ \\
    One turn         && $82$ & $\textbf{89}$ & $34$ && $\textbf{61}$ & $59$ & $12$ && $\textbf{95}$ & $90$ & $16$ && $\textbf{50}$   & \;$48$ & $20$ \\
    Navigation       && $80$ & $\textbf{86}$ & $14$ && $24$ & $\textbf{40}$ & $3$  && $\textbf{94}$ & $84$ & $2$  && $\textbf{47}$   & \;$44$ & $6$  \\
    Nav. dynamic     && $77$ & $\textbf{83}$ & $7$  && $24$ & $\textbf{38}$ & $2$  && $\textbf{89}$ & $82$ & $2$  && $\textbf{44}$   & \;$42$ & $4$  \\
\bottomrule
 \end{tabular}
}
\vspace{1mm}
\caption{Quantitative evaluation of three autonomous driving systems on goal-directed navigation tasks. The table reports the percentage of successfully completed episodes in each condition. Higher is better. The tested methods are: modular pipeline (MP), imitation learning (IL), and reinforcement learning (RL).}
\label{tbl:completion}
\end{table}

\mypara{Performance on the four tasks.}
Surprisingly, none of the methods performs perfectly even on the simplest task of driving straight on an empty street in the training conditions.
We believe the fundamental reason for this is variability in the sensory inputs encountered by the agents.
Training conditions include four different weather conditions.
The exact trajectories driven during training are not repeated during testing.
Therefore performing perfectly on this task requires robust generalization, which is challenging for existing deep learning methods.

On more advanced tasks the performance of all methods declines.
On the most difficult task of navigation in a populated urban environment, the two best methods~-- modular pipeline and imitation learning~-- are below $90\%$ success in all conditions and are below $45\%$ in the test town.
These results clearly indicate that performance is far from saturated even in the training conditions, and that generalization to new environments poses a serious challenge.

\mypara{Generalization.}
We study two types of generalization: to previously unseen weather conditions and to a previously unseen environment.
Interestingly, the results are dramatically different for these two.
For the modular pipeline and for imitation learning, the performance in the ``New weather'' condition is very close to performance in the training condition, and sometimes even better.
However, generalization to a new town presents a challenge for all three approaches.
On the two most challenging navigation tasks, the performance of all methods falls by at least a factor of $2$ when switching to the test town.
This phenomenon can be explained by the fact that the models have been trained in multiple weather conditions, but in a single town.
Training with diverse weather supports generalization to previously unseen weather, but not to a new town, which uses different textures and 3D models. The problem can likely be ameliorated by training in diverse environments.
Overall, our results highlight the importance of generalization for learning-based approaches to sensorimotor control.

\mypara{Modular pipeline vs end-to-end learning.}
It is instructive to analyze the relative performance of the modular pipeline and the imitation learning approach.
These systems represent two general approaches to designing intelligent agents, and CARLA enables a direct controlled comparison between them.

Surprisingly, the performance of both systems is very close under most testing conditions: the performance of the two methods typically differs by less than $10\%$.
There are two notable exceptions to this general rule.
One is that the modular pipeline performs better under the ``New weather'' condition than under the training conditions.
This is due to the specific selection of training and test weathers: the perception system happens to perform better on the test weathers.
Another difference between the two approaches is that MP underperforms on navigation in the ``New town'' condition and on going straight in ``New town\,\&\,weather''.
This is because the perception stack fails systematically under complex weather conditions in the context of a new environment.
If the perception stack is not able to reliably find a drivable path, the rules-based planner and the classic controller are  unable to navigate to the destination in a consistent way.
The performance is therefore bimodal: if the perception stack works, the whole system works well; otherwise it fails completely.
In this sense, MP is more fragile than the end-to-end method.

\mypara{Imitation learning vs reinforcement learning.}
We now contrast the performance of the two end-to-end trained systems: imitation learning and reinforcement learning.
On all tasks, the agent trained with reinforcement learning performs significantly worse than the one trained with imitation learning.
This is despite the fact that RL was trained using a significantly larger amount of data: $12$ days of driving, compared to $14$ hours used by imitation learning.
Why does RL underperform, despite strong results on tasks such as Atari games~\cite{Mnih2015dqn,Mnih2016a3c} and maze navigation~\cite{Mnih2016a3c,Dosovitskiy2017dfp}?
One reason is that RL is known to be brittle~\cite{Henderson2017}, and it is common to perform extensive task-specific hyperparameter search, such as $50$ trials per environment as reported by Mnih et al.~\cite{Mnih2016a3c}.
When using a realistic simulator, such extensive hyperparameter search becomes infeasible.
We selected hyperparameters based on evidence from the literature and exploratory experiments with maze navigation.
Another explanation is that urban driving is more difficult than most tasks previously addressed with RL.
For instance, compared to maze navigation, in a driving scenario the agent has to deal with vehicle dynamics and more complex visual perception in a cluttered dynamic environment.
Finally, the poor generalization of reinforcement learning may be explained by the fact that in contrast with imitation learning, RL has been trained without data augmentation or regularization such as dropout.

\begin{table}[]
\small
\centering
\ra{1.1}
\resizebox{1.0\linewidth}{!}{
 \begin{tabular}{@{}lcccccccccccccccc@{}}
\toprule

             && \multicolumn{3}{c}{Training conditions} && \multicolumn{3}{c}{New town} && \multicolumn{3}{c}{New weather} && \multicolumn{3}{c}{New town\,\&\,weather} \\
    Task                  && MP   & IL   & RL   && MP   & IL   & RL   && MP   & IL   & RL   && \;MP\; & \;IL   & \;RL\; \\
    \midrule
    Opposite lane         && $10.2$ & $\textbf{33.4}$ & $0.18$ && $0.45$ & $\textbf{1.12}$ & $0.23$ && $16.1$ & $\textbf{57.3}$ & $0.09$ &&  $0.40$  & \;$\textbf{0.78}$ & $0.21$  \\
    Sidewalk              && $\textbf{18.3}$ & $12.9$ & $0.75$ && $0.46$ & $\textbf{0.76}$ & $0.43$ && $24.2$ & $>\textbf{57}$ & $0.72$ &&  $0.43$  & \;$\textbf{0.81}$ & $0.48$  \\
    Collision-static      && $\textbf{10.0}$ & $5.38$ & $0.42$ && $\textbf{0.44}$ & $0.40$ & $0.23$ && $\textbf{16.1}$ & $4.05$ & $0.24$ &&  $\textbf{0.45}$  & \;$0.28$ & $0.25$  \\
    Collision-car         && $\textbf{16.4}$ & $3.26$ & $0.58$ && $0.51$ & $\textbf{0.59}$ & $0.41$ && $\textbf{20.2}$ & $1.86$ & $0.85$ &&  $\textbf{0.47}$  & \;$0.44$ & $0.37$  \\
    Collision-pedestrian  && $\textbf{18.9}$ & $6.35$ & $17.8$ && $1.40$ & $1.88$ & $\textbf{2.55}$ && $20.4$ & $11.2$ & $\textbf{20.6}$ &&  $1.46$  & \;$1.41$ & $\textbf{2.99}$  \\
\bottomrule
 \end{tabular}
}
\vspace{1mm}
\caption{Average distance (in kilometers) traveled between two infractions. Higher is better.}
\label{tbl:inf}
\vspace{-3mm}
\end{table}

\mypara{Infraction analysis.}
CARLA supports fine-grained analysis of driving policies. We now examine the behavior of the three systems on the hardest task: navigation in the presence of dynamic objects.
We characterize the approaches by average distance traveled between infractions of the following five types: driving on the opposite lane, driving on the sidewalk, colliding with other vehicles, colliding with pedestrians, and hitting static objects.
Details are provided in the supplement.

Table \ref{tbl:inf} reports the average distance (in kilometers) driven between two infractions.
All approaches perform better in the training town. For all conditions, IL strays onto the opposite lane least frequently, and RL is the worst in this metric. A similar pattern is observed with regards to veering onto the sidewalk.
Surprisingly, RL collides with pedestrians least often, which could be explained by the large negative reward incurred by such collisions.
However, the reinforcement learning agent is not successful at avoiding collisions with cars and static objects, while the modular pipeline generally performs best according to this measure.



These results highlight the susceptibility of end-to-end approaches to rare events: breaking or swerving to avoid a pedestrian is a rare occurrence during training. While CARLA can be used to increase the frequency of such events during training to support end-to-end approaches, deeper advances in learning algorithms and model architectures may be necessary for significant improvements in robustness~\cite{Chen:2015}.

\section{Conclusion}
\label{sec:conclusions}

We have presented CARLA, an open simulator for autonomous driving. In addition to open-source code and protocols, CARLA provides  digital assets that were created specifically for this purpose and can be reused freely. We leverage CARLA's simulation engine and content to test three approaches to autonomous driving: a classic modular pipeline, a deep network trained end-to-end via imitation learning, and a deep network trained via reinforcement learning. We challenged these systems to navigate urban environments in the presence of other vehicles and pedestrians. CARLA provided us with the tools to develop and train the systems and then evaluate them in controlled scenarios. The feedback provided by the simulator enables detailed analyses that highlight particular failure modes and opportunities for future work. We hope that CARLA will enable a broad community to actively engage in autonomous driving research. The simulator and accompanying assets will be released open-source at \url{http://carla.org}.

\section*{Acknowledgements}
CARLA would not have been possible without the development team at the CVC in Barcelona. The authors are particularly grateful to Nestor Subiron, the principal programmer, and Francisco Perez, the lead digital artist, for their tireless work. We sincerely thank artists Iris Saez and Alberto Abal, FX programmer Marc Garcia, and traffic behavior programmer Francisco Bosch.
We thank artists Mario Gonzalez, Juan Gonzalez, and Ignazio Acerenza for their contributions, and programmer Francisco Molero for his support. Antonio M. L\'{o}pez and Felipe Codevilla acknowledge the Spanish MINECO project TRA2014-57088-C2-1-R and the Spanish DGT project SPIP2017-02237, as well as the Generalitat de Catalunya CERCA Program and its ACCIO agency. Felipe Codevilla was supported in part by FI grant 2017FI-B1-00162. The authors thank Epic Games for support concerning the use of UE4.

\small
\bibliography{biblio}  

\normalsize

\newpage
\section*{Supplementary Material}

\renewcommand{\thesection}{S.\arabic{section}}
\setcounter{section}{0}
\renewcommand{\thefigure}{S.\arabic{figure}}
\setcounter{figure}{0}
\renewcommand{\thetable}{S.\arabic{table}}
\setcounter{table}{0}

\section{Simulator Technical Details}

\subsection{Client and Server Information Exchange}

CARLA is designed as a client-server system. The server runs and renders the CARLA world. The client provides an interface for users to interact with the simulator by controlling the agent vehicle and certain properties of the simulation.

\mypara{Commands.}
The agent vehicle is controlled by $5$ types of commands sent via the client:

\begin{itemize}
\item Steering: The steering wheel angle is represented by a real number between -1 and 1, where -1 and 1 correspond to full left and full right, respectively.
\item Throttle: The pressure on the throttle pedal, represented as a real number between 0 and 1.
\item Brake: The pressure on the brake pedal, represented as a real number between 0 and 1.
\item Hand Brake: A boolean value indicating whether the hand brake is activated or not.
\item Reverse Gear: A boolean value indicating whether the reverse gear is activated or not.
\end{itemize}

\mypara{Meta-commands.}
The client is also able to control the environment and the behavior of the server with the following meta commands:
\begin{itemize}
\item Number of Vehicles: An integer number of non-player vehicles to be spawned in the city.
\item Number of Pedestrians: An integer number of pedestrians to be spawned in the city.
\item Weather Id: An index of the weather/lighting presets to use. The following are currently supported:
Clear Midday, Clear Sunset, Cloudy Midday, Cloudy Sunset, Soft Rain Midday, Soft Rain Sunset, Medium Rain Midday, Cloudy After Rain Midday, Cloudy After Rain Sunset,
Medium Rain Sunset,  Hard Rain Midday, Hard Rain Sunset, After Rain Noon, After Rain Sunset.
\item Seed Vehicles/Pedestrians: A seed that controls how non-player vehicles and pedestrians are spawned. It is possible to have the same vehicle/pedestrian behavior by setting the same seed.
\item Set of Cameras: A set of cameras with specific parameters such as position, orientation, field of view, resolution and camera type. Available camera types include an optical RGB camera, and pseudo-cameras that provide ground-truth depth and semantic segmentation.
\end{itemize}

\mypara{Measurements and sensor readings.}
The client receives from the server the following information about the world and the player's state:
\begin{itemize}
\item Player Position: The 3D position of the player with respect to the world coordinate system.
\item Player Speed: The player's linear speed in kilometers per hour.
\item Collision: Cumulative impact from collisions with three different types of objects: cars, pedestrians, or static objects.
\item Opposite Lane Intersection: The current fraction of the player car footprint that overlaps the opposite lane.
\item Sidewalk Intersection: The current fraction of the player car footprint that overlaps the sidewalk.
\item Time: The current in-game time.
\item Player Acceleration: A 3D vector with the agent's acceleration with respect to the world coordinate system.
\item Player Orientation: A unit-length vector corresponding to the agent car orientation.
\item Sensor readings: The current readings from the set of camera sensors.
\item Non-Client-Controlled agents information: The positions, orientations and bounding boxes for all pedestrians
and cars present in the environment.
\item Traffic Lights information: The
position and state of all traffic lights.
\item Speed Limit Signs information:
Position and readings from all speed limit signs.
\end{itemize}

\subsection{Environment}

\begin{figure}
	\centering
  \setlength{\tabcolsep}{3pt}
		\begin{tabular}{cc}
			\includegraphics[width=6.5cm]{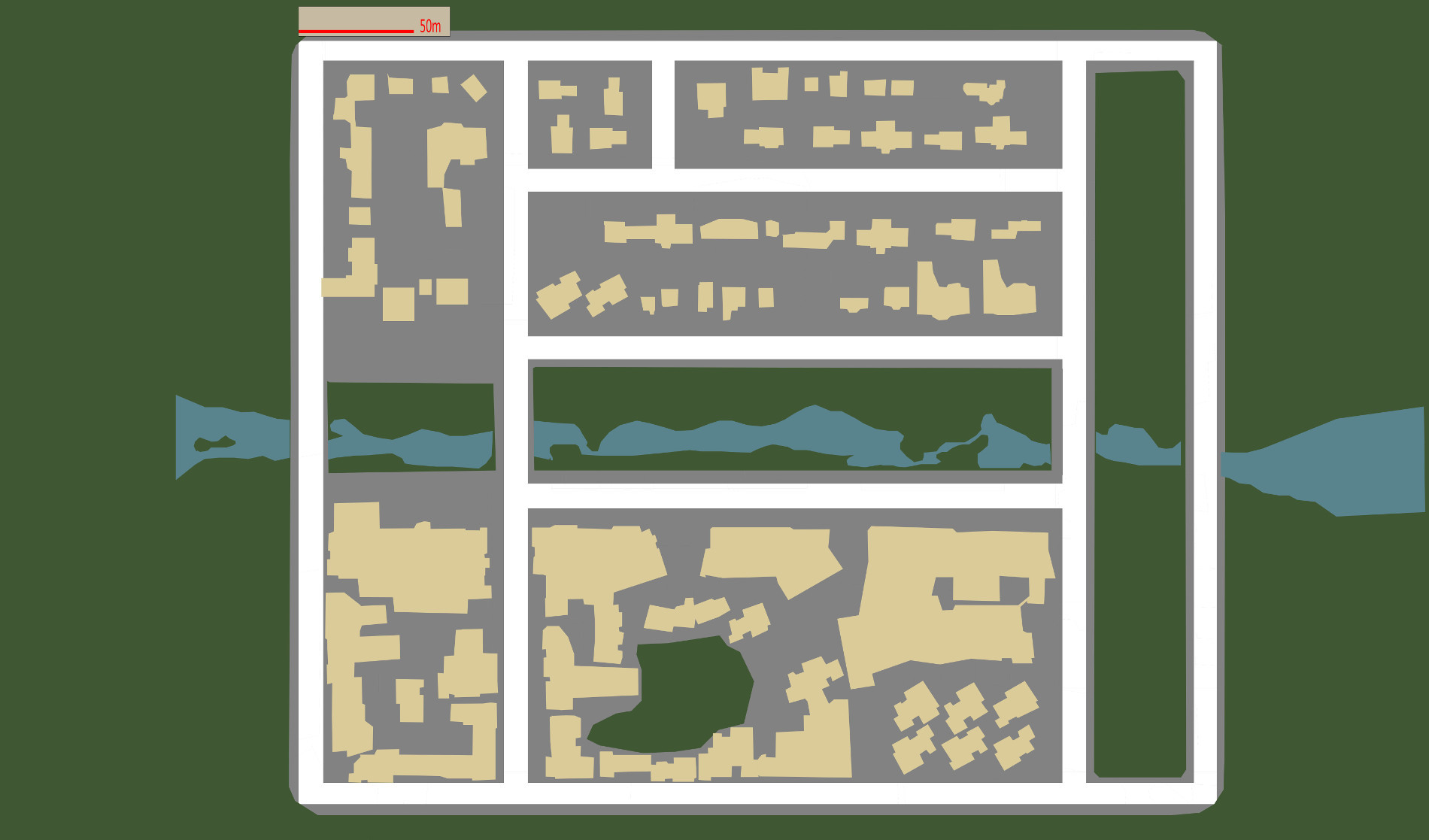} &
			\includegraphics[width=6.5cm]{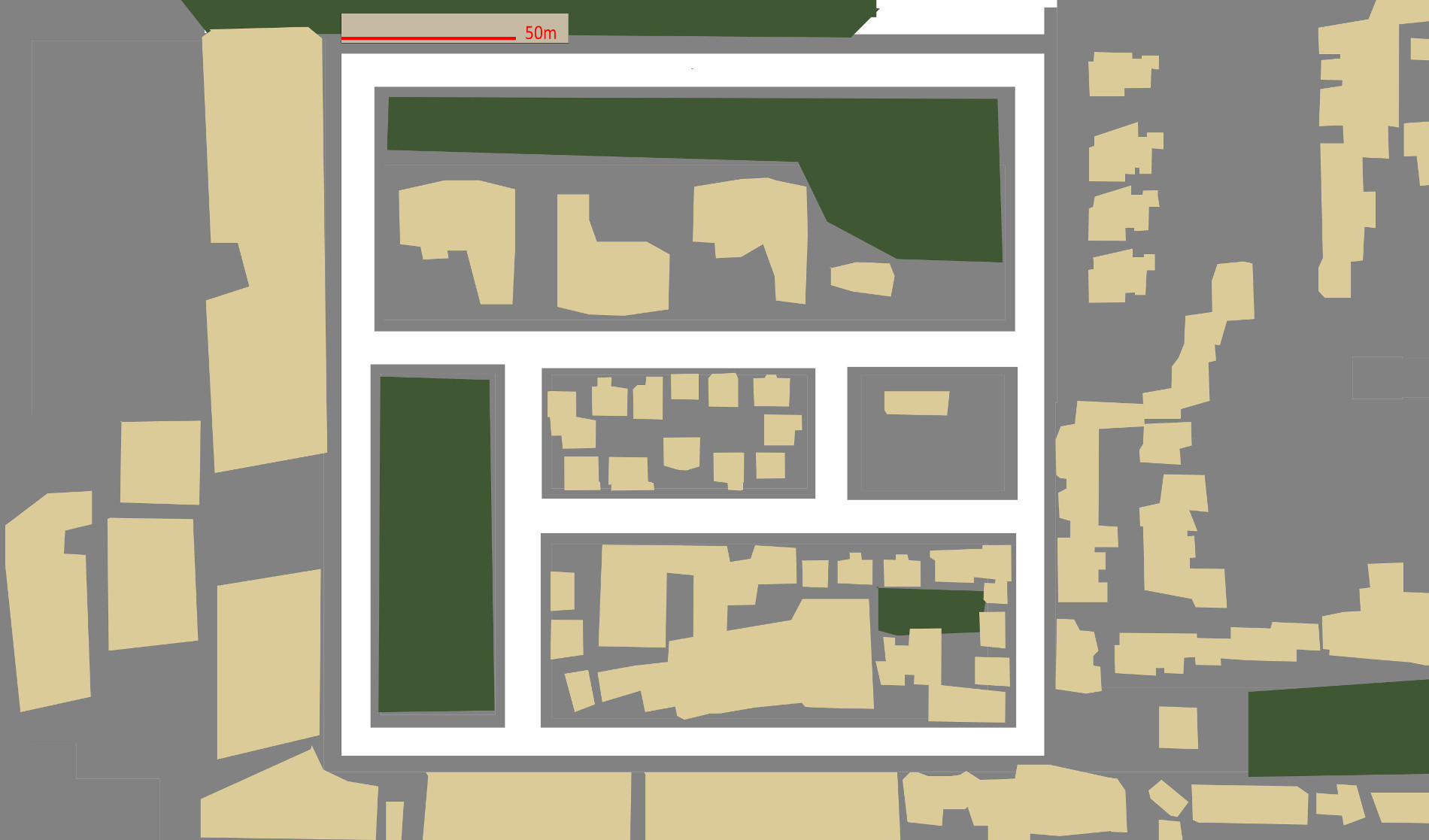}\\
			  Town 1 Map  &  Town 2 Map	  \\

			\includegraphics[width=6.5cm]{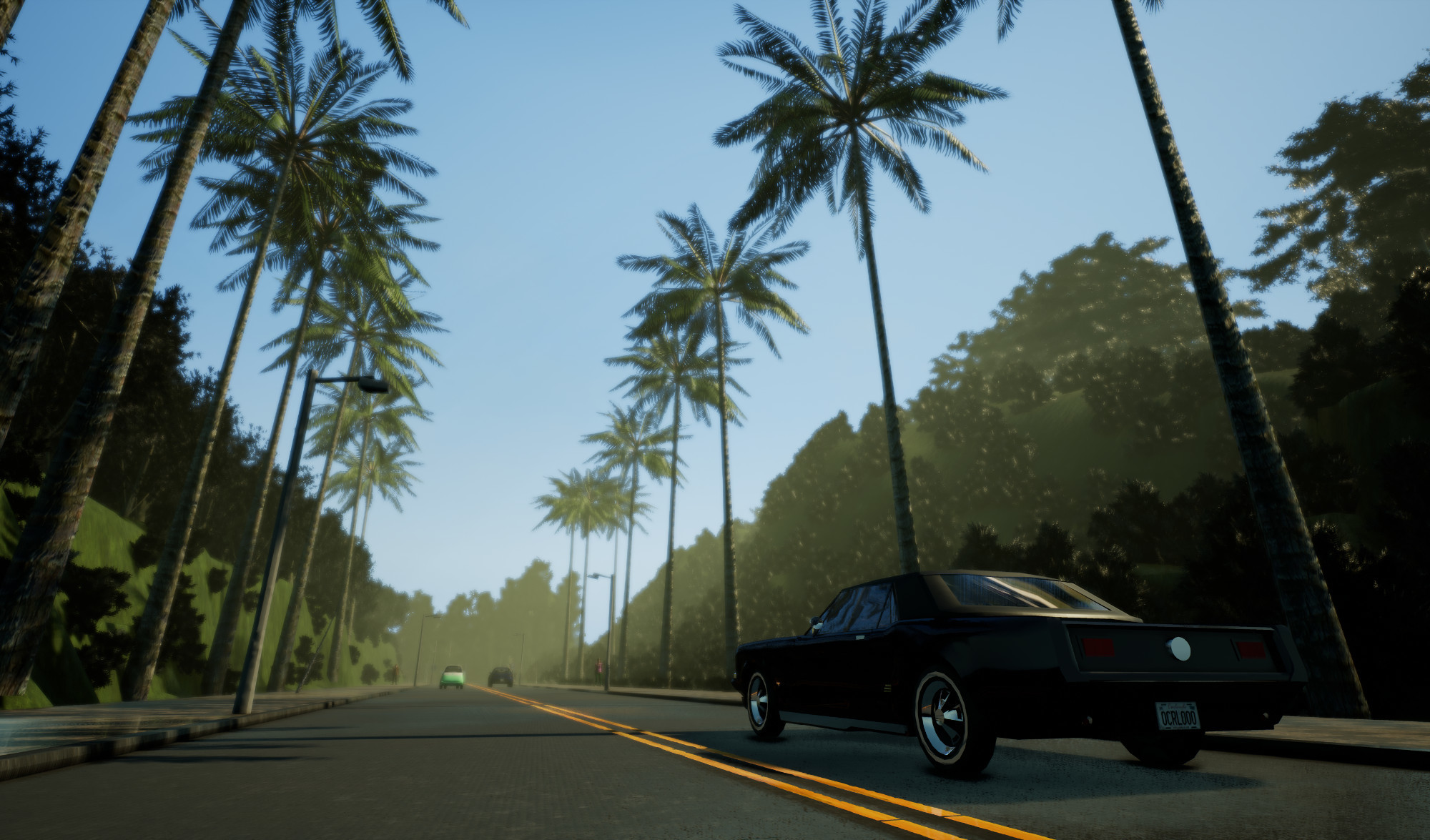}&
			\includegraphics[width=6.5cm]{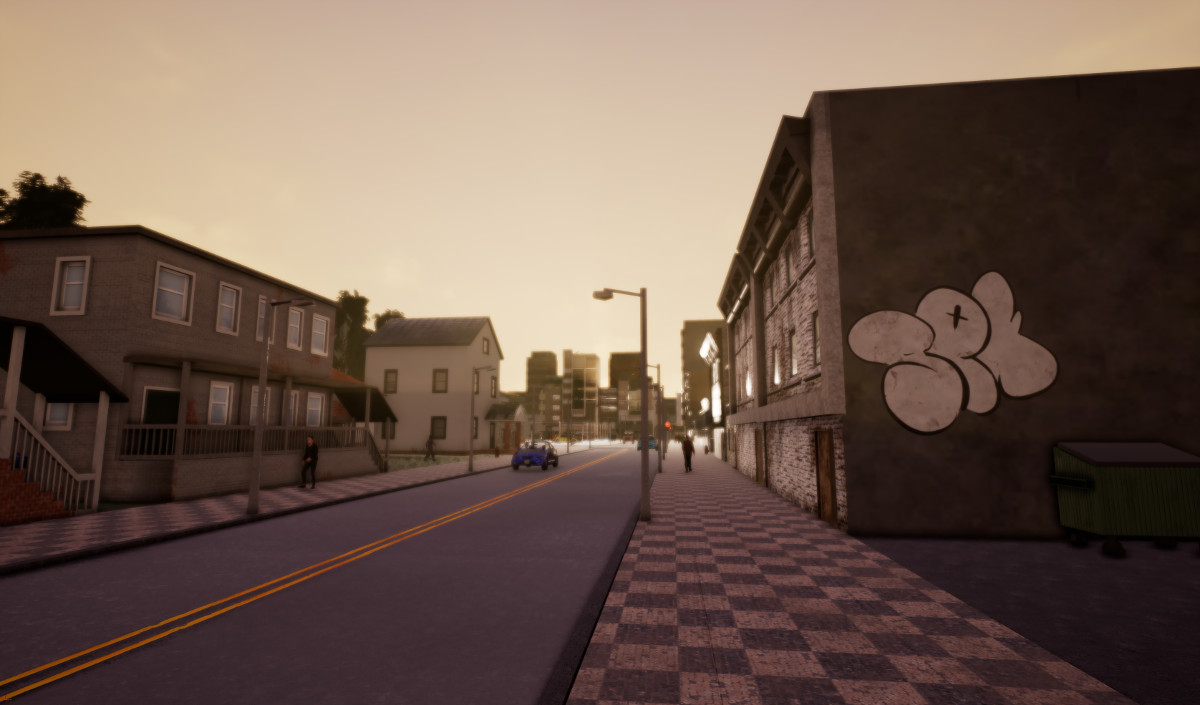}\\
			\includegraphics[width=6.5cm]{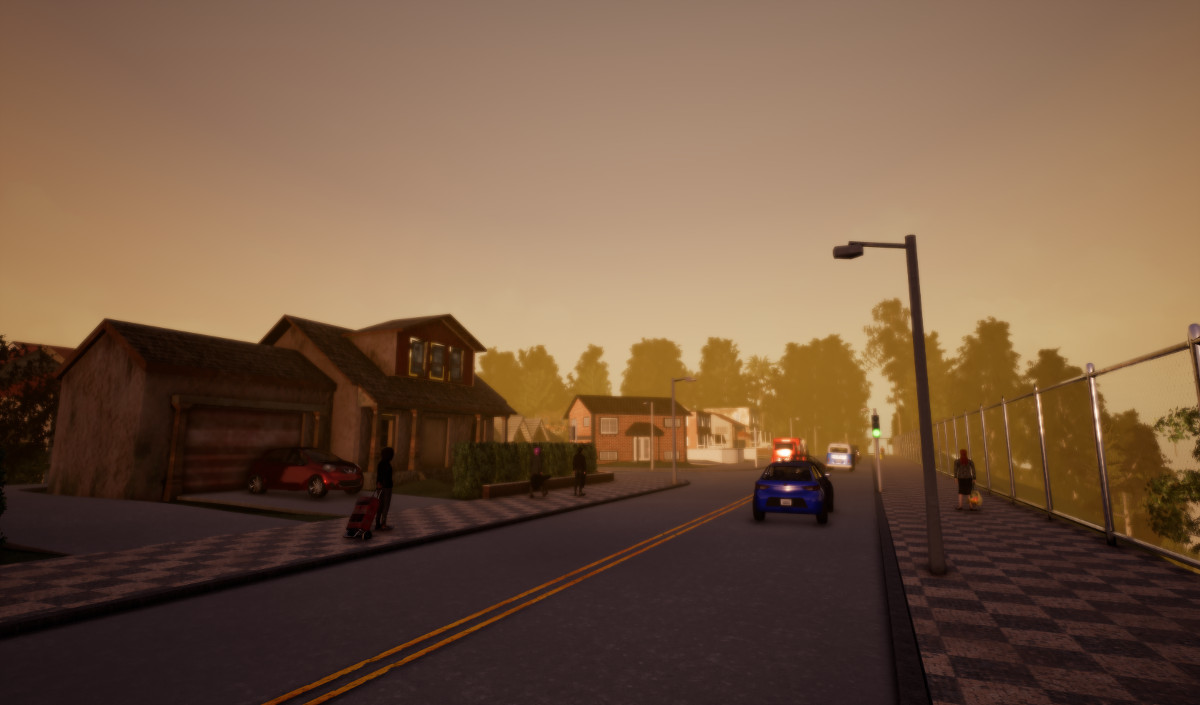} &
			\includegraphics[width=6.5cm]{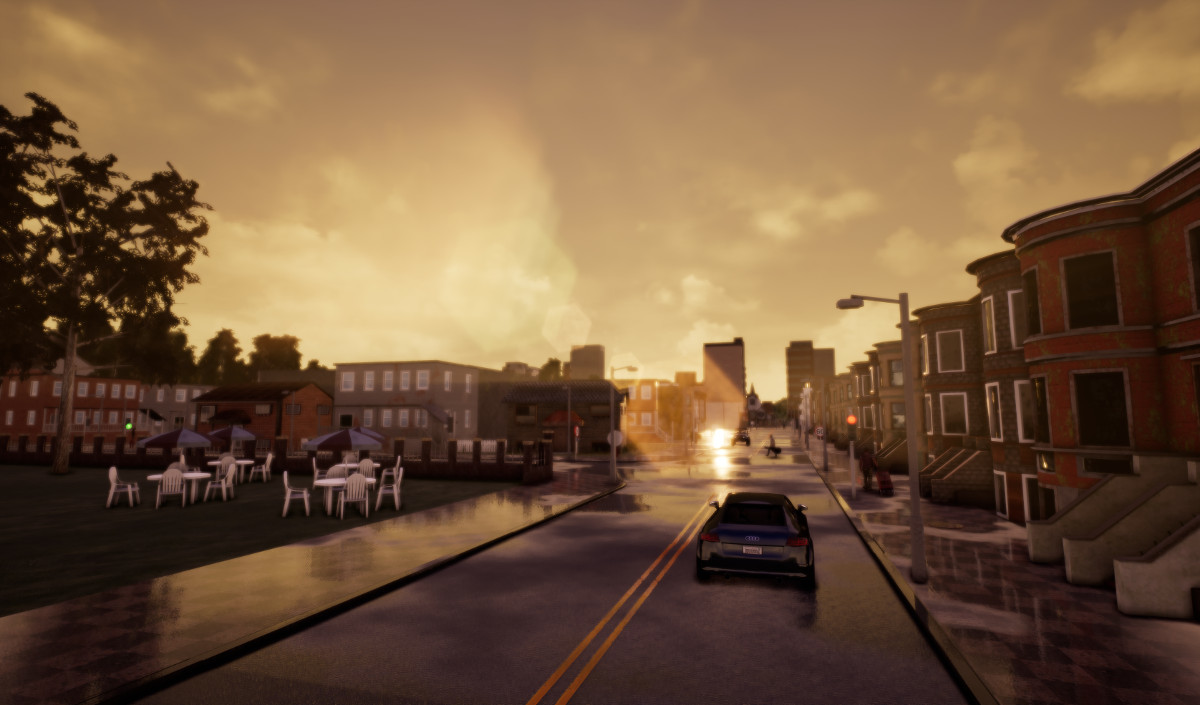}\\

			\includegraphics[width=6.5cm]{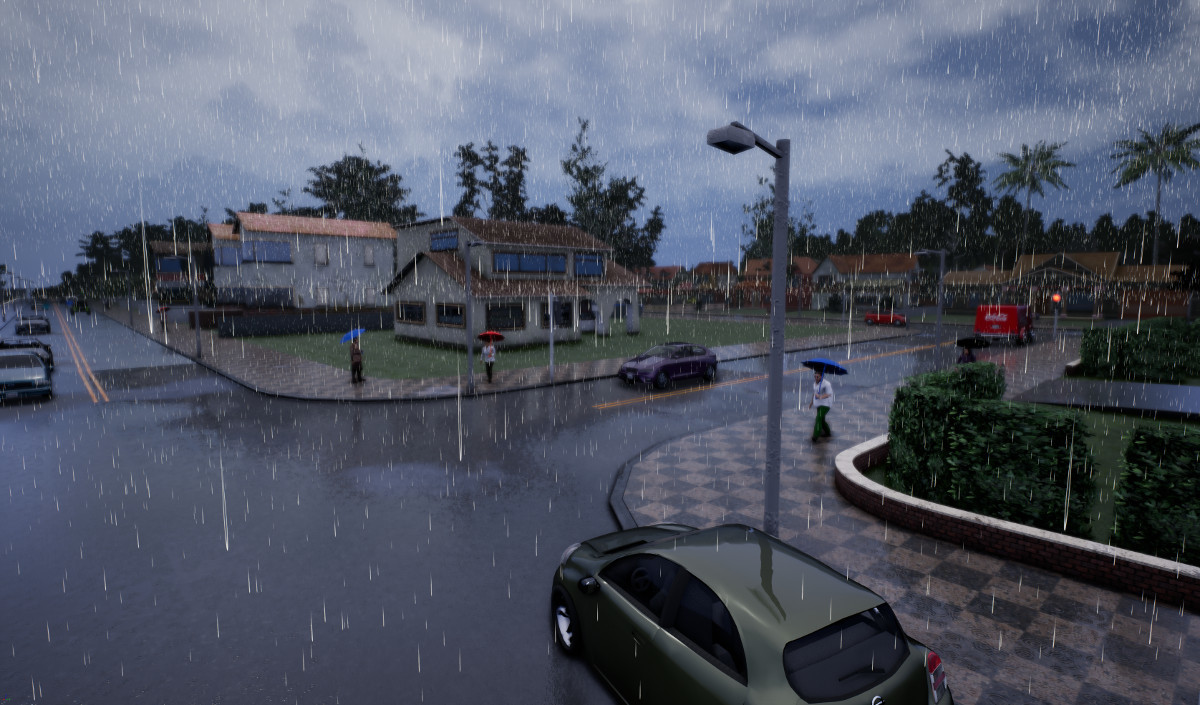} &
			\includegraphics[width=6.5cm]{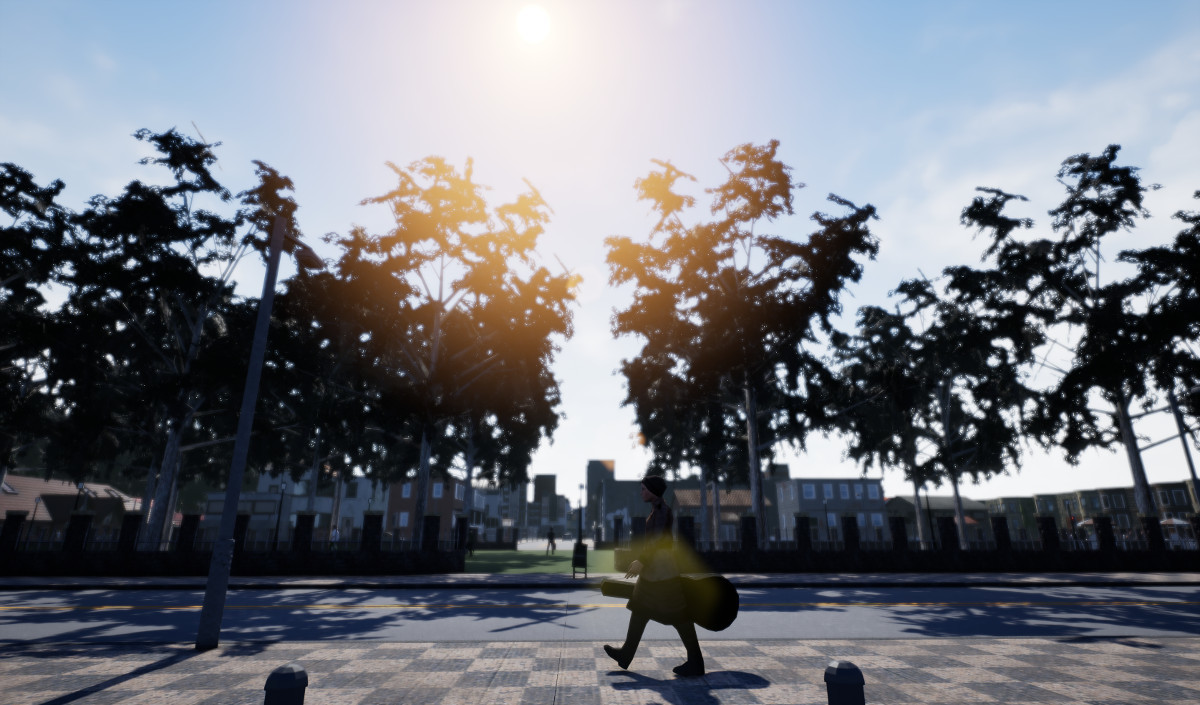}\\
			\includegraphics[width=6.5cm]{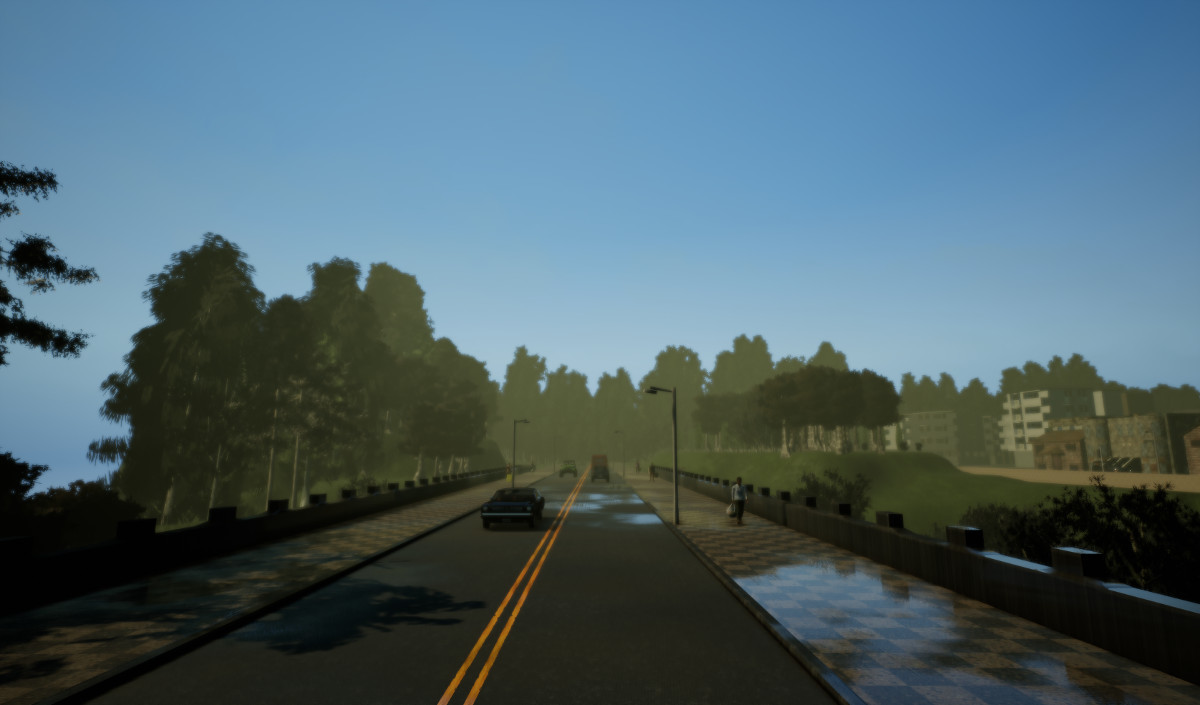} &
			\includegraphics[width=6.5cm]{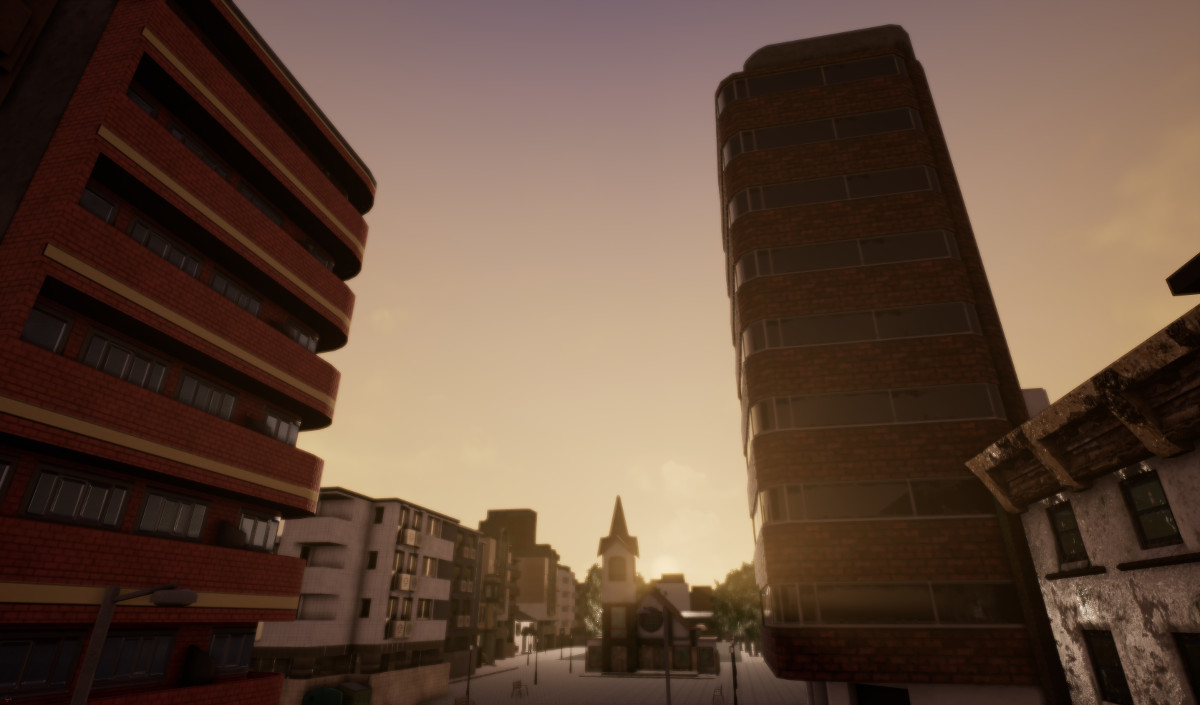}

		\end{tabular}
    \caption{The two CARLA towns. \textbf{Left:} views and a map of CARLA Town 1. \textbf{Right:} views and a map of CARLA Town 2.}
	  \label{fig:towns}

\end{figure}

\begin{figure}
	\centering
  \setlength{\tabcolsep}{3pt}
		\begin{tabular}{c}
			\includegraphics[scale=0.16]{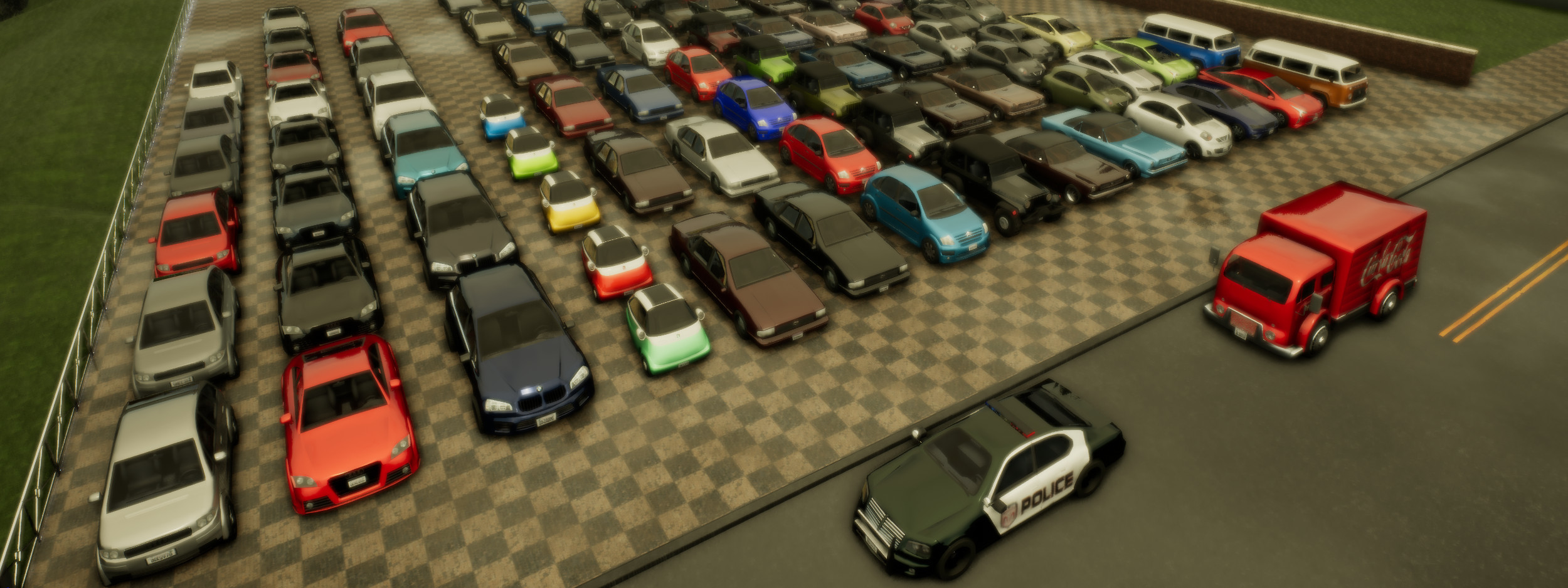} \\
			\includegraphics[scale=0.16]{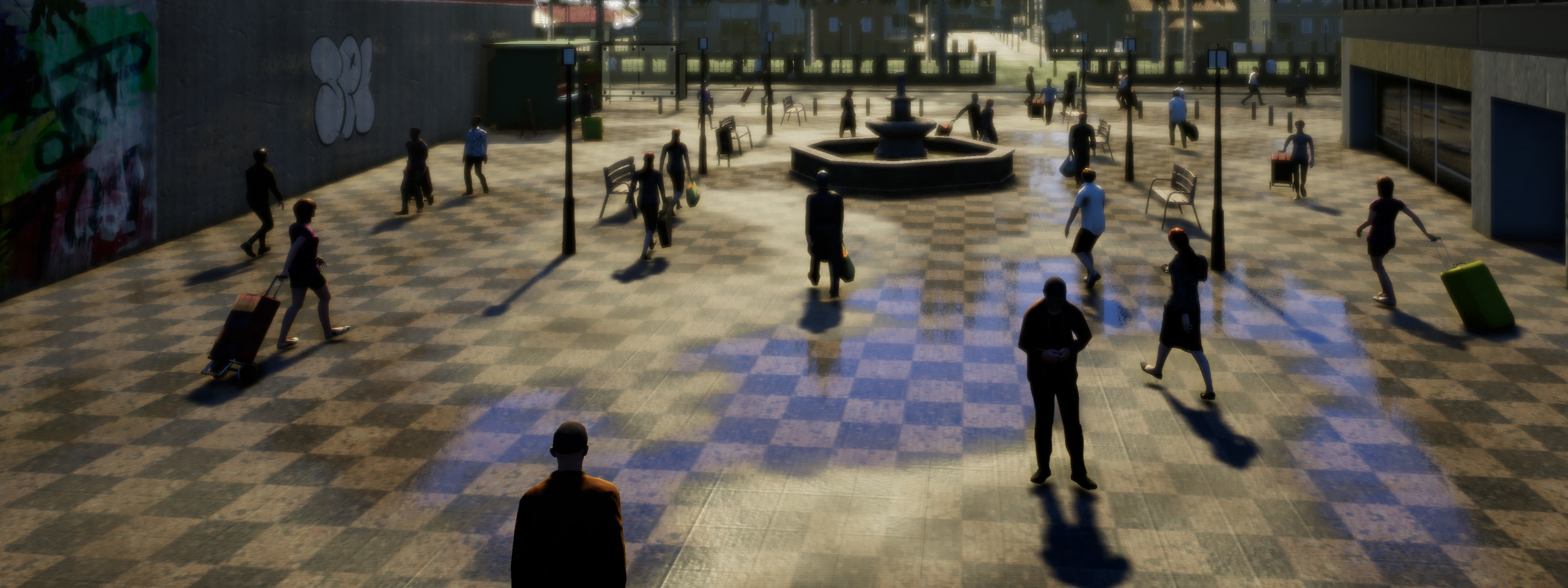}\\
 			\includegraphics[scale=0.16]{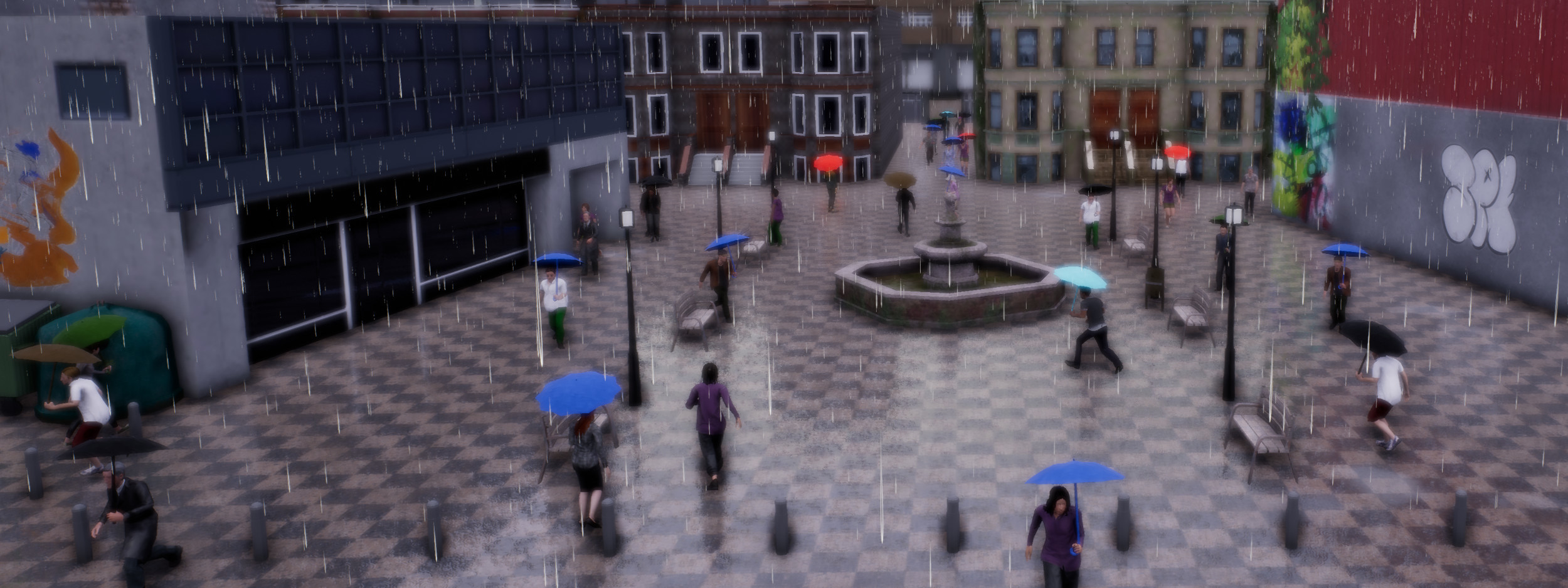}
		\end{tabular}
    \caption{Diversity of cars and pedestrians currently available in CARLA.}
	  \label{fig:carandpeds}
\end{figure}

CARLA provides two towns: Town 1 and Town 2. Figure \ref{fig:towns} shows maps of these towns and representative views.
A large variety of assets were produced for CARLA, including cars and pedestrians. Figure \ref{fig:carandpeds} demonstrates this diversity.

\section{Driving Systems Technical Details}

In this section we provide additional technical details for the autonomous driving systems we have tested.

\subsection{Modular Pipeline}

\mypara{Perception module.}
Training of the semantic segmentation network was performed using Adam~\cite{Kingma2015adam} with learning rate $\lambda_0 = 10^{-6}$ for $300$ epochs with batch size $2$.
The back-end ResNet is pre-trained on ImageNet and frozen during training.
No data augmentation is used.

The intersection classifier network is trained on $500$ images balanced between the two classes.
We used Adam with learning rate $\lambda_0 = 10^{-3}$ for $500$ epochs with batch size $5$. No pre-training or data augmentation are used.

\mypara{Local planning for modular pipeline.}
In the \texttt{road-following} state, the local planner uses the ego-lane mask computed from the semantic segmentation to select points that maintain a fixed distance with the right edge of the road. The \texttt{left-turn} at intersections is more complex due to the temporary absence of lane markings, the longer distance to the target lane, and the limited field of view of the forward-facing camera. To deal with these challenges, we first compute waypoints towards the center of the intersection with a predefined skew angle; this helps improve the visibility of the target lane. An auxiliary camera (along with its computed semantic segmentation map) is used to determine the shape and alignment of the vehicle with respect of the target lane. In the second step, the waypoints are laid out to yield a smooth trajectory from the center of the intersection to the target lane. The \texttt{right-turn} state uses a similar strategy. However, as turning right is easier given the target lane proximity, the number of waypoints needed is lower and only the forward-facing information is required. The \texttt{intersection-forward} state is handled similarly to road-following.
The \texttt{hazard-stop} mode is activated when the dynamic obstacle map presents a cumulative probability of obstacle above a predefined threshold. In this case, the system generates a special waypoint to request an emergency break from the controller.

\subsection{Imitation Learning}

\mypara{Architecture.}
Table~\ref{tbl:arch} details the configuration of the network used in the imitation learning approach~\cite{Codevilla2017}.
The network is composed of four modules: a perception module that is focused on processing image inputs, a measurement module that processes the speed input, a joint input module that merges the perception and measurement information, and a control module that produces motor commands from the joint input representation.
The control module consists of $4$ identical branches: command-conditional modules for predicting the steering angle, brake, and throttle -- one for each of the four commands.
One of the four command-conditional modules is selected based on the input command.
The perception module is implemented by a convolutional network that takes a $200 \timess 88$ image as input and outputs a $512$-dimensional vector.
All other modules are implemented by fully-connected networks.
The measurement module takes as input a measurement vector and outputs a $128$-dimensional vector.

\begin{table}[]
  \centering
    \begin{tabular}{|c|cccc|}
      \hline
      module                       & input dimension           & channels            & kernel & stride \\ \midrule
      \multirow{10}{*}{Perception}  & $200 \times 88 \times 3$   & $32$                & $5$    & $2$    \\
								  & $98 \times 48 \times 32$      & $32$                & $3$    & $1$    \\
                                   & $96 \times 46 \times 32$  & $64$                & $3$    & $2$    \\
								  & $47 \times 22 \times 64$  & $64$                & $3$    & $1$    \\
                                   & $45 \times 20 \times 64$  & $128$                & $3$    & $2$    \\
								  & $22 \times 9 \times 128$  & $128$                & $3$    & $1$    \\
                                   & $20 \times 7 \times 128$   & $256$               & $3$    & $2$   \\
                                   & $9 \times 3 \times 256$   & $256$               & $3$    & $1$   \\
                                   & $7 \cdot 1 \cdot 256$   & $512$               & $-$    & $-$   \\

                                   & $512$   & $512$               & $-$    & $-$

\\ \hline
      \multirow{3}{*}{Measurement} & $1$                       & $128$               & $-$    & $-$    \\
                                   & $128$                     & $128$               & $-$    & $-$    \\
                                   & $128$                     & $128$               & $-$    & $-$    \\  \hline
      \multirow{1}{*}{Joint input} & $512 + 128$               & $512$       & $-$    & $-$    \\ \hline
      \multirow{3}{*}{Control}     & $512$                     & $256$       & $-$    & $-$    \\
									                 & $256$                     & $256$         & $-$    & $-$      \\
                                   & $256$                     & $1$         & $-$    & $-$      \\ \hline
    \end{tabular}

    \vspace{0.5cm}
  \caption{Exact configurations of all network modules for the imitation learning approach.}
  \label{tbl:arch}
\vspace{-3mm}
\end{table}

\mypara{Training details.}
We trained all networks with Adam~\cite{Kingma2015adam}.
We used mini-batches of $120$ samples.
We balanced the mini-batches, using the same number of samples for each command.
Our starting learning rate was $0.0002$ and it was multiplied by $0.5$ every $50@000$ mini-batch iterations.
We trained for $294@000$ iterations in total.
Momentum parameters were set to $\beta_1 = 0.7$ and $\beta_2 = 0.85$.
We used no weight decay, but performed $50\%$ dropout after hidden fully-connected layers and $20\%$
dropout on convolutional layers.
To further reduce overfitting, we performed extensive data augmentation by adding Gaussian blur, additive Gaussian noise, pixel dropout, additive and multiplicative brightness variation, contrast variation, and saturation variation.
Before feeding a raw $800 \times 600$ image to the network, we cropped $171$ pixels at the top and $45$ at the bottom, and then resized the resulting $800 \times 384$ image to a resolution of $200 \timess 88$.

\mypara{Training data.}
The expert training data was collected from two sources: an automated agent and human driver data.
The automated agent has access to privileged information such as locations of dynamic objects, ego-lane, states of traffic lights. $80\%$ of the demonstrations were provided by the automated agent and $20\%$ by a human driver.

In order to improve the robustness of the learned policy, we injected noise into the expert's steering during training data collection. Namely, at random points in time we added a perturbation to the steering angle provided by the driver.
The perturbation is a triangular impulse: it increases linearly, reaches a maximal value, and then linearly declines.
This simulates gradual drift from the desired trajectory, similar to what might happen with a poorly trained controller.
The triangular impulse is parametrized by its starting time $t_0$, duration $\tau \in \Re^+$, sign $\sigma \in \{-1,\, +1\}$, and intensity $\gamma \in \Re^+$:
\begin{equation}
  s_{perturb}(t) = \sigma \gamma \max \left(0, \left(1 - \left|\frac{2(t-t_0)}{\tau} - 1 \right|\right)\right).
\end{equation}
Every second of driving we started a perturbation with probability $p_{perturb}$.
We used $p_{perturb} = 0.1$ in our experiments.
The sign of each perturbation was sampled at random, the duration was sampled uniformly from $0.5$ to $2$ seconds, and intensity was fixed to $0.15$.

\subsection{Reinforcement Learning}
We base our A3C agent on the network architecture proposed by Mnih et al.~\cite{Mnih2015dqn}.
The input to the network consists of two most recent images observed by the agent, resized to $84 \times 84$ pixels, as well as a vector of measurements.
The measurement vector includes the current speed of the car, distance to goal, damage from collisions, and the current high-level command provided by the topological planner, in one-hot encoding.
The inputs are processed by two separate modules: images by a convolutional module, measurements by a fully-connected network.
The outputs of the two modules are concatenated and further processed jointly.
We trained A3C with $10$ parallel actor threads, for a total of $10$ million environment steps.
We used $20$-step rollouts, following Jaderberg et al.~\cite{Jaderberg2017unreal}, with initial learning rate $0.0007$ and entropy regularization $0.01$.
Learning rate was linearly decreased to zero over the course of training.

The reward is a weighted sum of five terms: distance traveled towards the goal $d$ in km, speed $v$ in km/h, collision damage $c$, intersection with the sidewalk $s$ (between $0$ and $1$), and intersection with the opposite lane $o$ (between $0$ and $1$).
\begin{equation}
	r_t = 1000\, (d_{t-1} - d_{t}) + 0.05\, (v_t - v_{t-1})  - 0.00002\, (c_t - c_{t-1}) - 2\, (s_t - s_{t-1}) - 2\, (o_t - o_{t-1}).
\end{equation}

\section{Experimental Setup}

\subsection{Types of Infractions}

We characterize the approaches by average distance travelled between infractions of the following five types:
\begin{itemize}
\setlength{\itemsep}{0pt}
\setlength{\parskip}{0pt}
\setlength{\parsep}{0pt}
\item Opposite lane: More than $30\%$ of the car's footprint is over wrong-way lanes.
\item Sidewalk: More than $30\%$ of the car's footprint is over the sidewalk.
\item Collision with static object: Car makes contact with a static object, such as pole or building.
\item Collision with car: Car makes contact with another car.
\item Collision with pedestrian: Car makes contact with a pedestrian.
\end{itemize}
The duration of each violation is limited to $2$ seconds, so if the car remains on the sidewalk for $10$ seconds, this will be counted as $5$ violations, not one.

\end{document}